\documentclass[preprint,12pt]{elsarticle}

\usepackage{amssymb}
\usepackage{amsmath}
\usepackage{multirow}
\usepackage{tabularx}
\usepackage{rotating}
\usepackage{subcaption}
\usepackage{array}
\usepackage{threeparttable}
\usepackage{url}

\journal{Acta Astronautica}

\begin{document}

\begin{frontmatter}



\title{Addressing Camera Sensors Faults in Vision-Based Navigation: Simulation and Dataset Development} %% Article title


\author[label1,label2]{Riccardo Gallon} %% Author name
\ead{r.gallon@tudelft.nl}

\author[label2]{Fabian Schiemenz}
% \author[inst2]{Alisa \snm{Krstova}}
\author[label1]{Alessandra Menicucci}
\author[label1]{Eberhard Gill}

\affiliation[label1]{organization={Department of Space Systems Engineering, Faculty of Aerospace Engineering},%Department and Organization
            addressline={Kluyverweg 1}, 
            city={HS Delft},
            postcode={2629},
            country={Netherlands}}

\affiliation[label2]{organization={Airbus Defence and Space GmbH},%Department and Organization
            addressline={Claude-Dornier Stra\ss e}, 
            city={Immenstaad am Bodensee},
            postcode={88090}, 
            country={Germany}}

%% Abstract
\begin{abstract}
The increasing importance of Vision-Based Navigation (VBN) algorithms in space missions raises numerous challenges in ensuring their reliability and operational robustness. Sensor faults can lead to inaccurate outputs from navigation algorithms or even complete data processing faults, potentially compromising mission objectives. Artificial Intelligence (AI) offers a powerful solution for detecting such faults, overcoming many of the limitations associated with traditional fault detection methods. However, the primary obstacle to the adoption of AI in this context is the lack of sufficient and representative datasets containing faulty image data.

This study addresses these challenges by focusing on an interplanetary exploration mission scenario. A comprehensive analysis of potential fault cases in camera sensors used within the VBN pipeline is presented. The causes and effects of these faults are systematically characterized, including their impact on image quality and navigation algorithm performance, as well as commonly employed mitigation strategies. To support this analysis, a simulation framework is introduced to recreate faulty conditions in synthetically generated images, enabling a systematic and controlled reproduction of faulty data. The resulting dataset of fault-injected images provides a valuable tool for training and testing AI-based fault detection algorithms. The final link to the dataset will be added after an embargo period. For peer-reviewers, this private link\footnote{The link will be available upon publication. For the time being, readers are kindly requested to contact the corresponding author to obtain access to the dataset.}
%\url{https://data.4tu.nl/private_datasets/zhYLatf4QrcwrFzUGuZi9j9OSLZ6a_9q4D2lINsR5ZM}}
is available.
\end{abstract}


% \begin{highlights}
% \end{highlights}

%% Keywords

\end{frontmatter}



\section{Introduction}\label{sec1}
%% Labels are used to cross-reference an item using \ref command.

The most promising trends in recent space exploration target ambitious technological innovations and advanced operations, such as Lunar and Martian exploration, in-orbit refuelling and maintenance, and active debris removal. These complex endeavours heavily rely on vision-based navigation (VBN), marking a significant departure from traditional satellite operations. As we venture into these uncharted territories, the reliability and accuracy of visual sensors, including LiDARs and cameras, become paramount to mission success.

In the present work, framed within the project \textit{Astrone KI} \citep{martin2023pioneering}, the VBN approach is used to perform autonomous relocation in a harsh Small Solar System Body (SSSB) environment, leveraging AI to enhance the VBN task \citep{olucak2023sensor}, \citep{liesch2023ai}. Additionally, the project aims for an AI-augmented FDIR subsystem \citep{gallon2024machine}, \citep{gallon2024convolutional}, in order to manage potential faults arising in onboard sensors and maintain nominal VBN operations. Since FDIR based on the traditional Packet Utilization Standard (PUS) \citep{ECSS-E-ST-70-41C} struggles with the multi-dimensional nature of image data, AI has been proposed as a promising solution for VBN-based FDIR, thanks to its historically-proven effectiveness in dealing with image processing tasks.

However, the peculiarities of the VBN approach require the present analysis to investigate on the nature and source of the possible issues occurring in visual sensors. Depending on the application, different factors can affect commonly employed cameras and LiDARs, such as dust covering the optics after landing, malfunctioning pixels, light reflections, etc. A precise assessment of these faults, up to a proper Failure Modes and Effects Analysis (FMEA, \citep{ECSS-Q-ST-30-02C}), will mark a key step towards the reliability of VBN systems. Nevertheless, the development of an accurate simulation environment, inclusive of the aforementioned failure modes, is crucial to the integration of AI into spaceflight solutions. The ability to realistically model and generate diverse fault scenarios in visual data is essential for training and validating AI algorithms, ensuring their effectiveness and reliability in the FDIR task.

This paper presents a novel approach to simulate faults affecting optical sensors which enable the vision-based navigation subsystem of \textit{Astrone KI}, laying the groundwork for the development of the AI-powered FDIR subsystem. By addressing this crucial gap in the development streamline, this work takes initial steps towards the adoption of AI technologies in space applications, whose ultimate goal should be to enable safer, more efficient, and more ambitious missions in our solar system and beyond. Finally, a dataset which makes use of the mentioned faults simulation is proposed, providing a useful tool for training and testing AI algorithms to accomplish the FDIR task.

The structure of this paper is organized as follows. Section \ref{sec:rel_work} presents relevant work in the field of camera sensor faults simulation and datasets generation for VBN applications. Section \ref{sec:envdesc_selfail} introduces the simulation environment and the considered fault cases, detailing the issues each fault creates to the camera, the effects causing the fault, its manifestation in the final image and any typically adopted mitigation strategies, if available. Section \ref{sec:datagen} outlines the dataset generations strategy, including the image acquisition methodology, the faults injection process, and the generation of label masks. Finally, Section \ref{sec:concl_fw} concludes the paper by summarizing the main achievements and providing an outlook on potential future developments.

\section{Related Work}\label{sec:rel_work}

% \begin{itemize}
%     \item Datasets of Images in the Space Domain
%     \item Datasets for FDIR
%     \item Table comparison of the different available datasets in terms of certain criteria pertinent to the problem: label availability, image resolution, image count, license type (open source vs proprietary), ...
% \end{itemize}

Since the introduction of AI-based VBN solutions in the space domain with the Mars Exploration Rover (MER) mission \citep{maimone2006autonomous}, the significance of data for the training and testing of algorithms has become increasingly evident. The requirements of having realistic and representative images has led researchers to address the topic of dataset creation to support their AI developments \citep{sharma2019speed}, \citep{park2022speed+}, \citep{lebreton2024training}. Datasets for the specific case of landing and navigation on extra-terrestrial bodies are very limited, primarily because of the reduced number of missions targeting the mentioned scenarios, but also because of the challenges of \textit{labelling}. The process of labelling consists in assigning an image or each of its pixels to a specific class, which will be employed during the AI training to learn what the specific image or pixel represents. Typical examples of classes used in VBN can be landmarks (e.g., rocks, crater), \textit{Sun}, \textit{sky} and \textit{distant landscape}, but also \textit{solar panel}, \textit{satellite body}, \textit{docking port}, depending on the specific VBN task.

While missions such as Hayabusa, Hayabusa 2 and Osiris-REx pioneered the exploration of SSSBs and retrieved a gigantic number of images and scientific data during their respective landing phases with their camera sensors \citep{bos2018touch}, \citep{yamada2023inflight}, the effort of labelling every single image or even pixel to be able to use those images for training navigation tasks has not yet received an efficient solution, thus leaving a significant gap in the state of the art. The problem of labelling is far from being solved. Thus, researchers are mostly oriented towards the generation of synthetic datasets which provide a faithful representation of the environment and automatically output labels accordingly. Some synthetic datasets exist in literature to accomplish navigation tasks, including satellite pose estimation, in-orbit rendezvous and landing \citep{sharma2019speed}, \citep{park2022speed+}, \citep{lebreton2024training}. \citet{sharma2019speed} and \citet{park2022speed+} are pioneering works in the field of spacecraft pose estimation, dealing with the problem of navigation around non-cooperative space objects and their relative position estimation. \citet{sharma2019speed} proposes a synthetic dataset comprehensive of camera images of an actual spacecraft mock-up, which are fused to simulated images of the Earth and the Sun to enhance the realism of the result. For the same pose estimation task, \citet{park2022speed+} proposes an improvement of the SPEED dataset \citep{sharma2019speed}, SPEED+, which comprises fully synthetic images and Hardware-in-the-Loop test images taken exploiting a robotic simulation testbed. \citet{lebreton2024training} addresses the task of pose estimation in two different scenarios, one is relative pose estimation during rendezvous of non-cooperative objects and the other is self pose estimation during Moon landing. In \citet{lebreton2024training}, a number of different datasets belonging to the two scenarios is proposed. The synthetic datasets are entirely generated with a high-fidelity image simulator (SurRender, \citep{brochard2018scientific}), employing digitalized models of mock-ups of the Envisat satellite. The same mock-up is used to obtain completely experimental images exploiting the TRON facility at DLR \citep{kruger2010tron}, which was already utilized in \citet{park2022speed+}. Finally Generative Adversarial Networks (GANs) are leveraged to reproduce the effects of real camera acquisition on the dataset images, trying to bridge the domain gap that real images have with respect to synthetic ones.\\
As \citet{lebreton2024training} is the only attempt reported in the state of the art to obtain a realistic dataset for landing applications, it does not cover the exploration of extra-terrestrial bodies, as in the case of \textit{Astrone KI}. However, the similarities between the terrain of the Moon and any other Small Solar System Body could support the use of a dataset generated on the former in training and testing an AI for navigating on the latter, but no work as been done so far in this field. Recently, NASA has been releasing labelled data from the Opportunity and Spirit rovers from NASA's Mars Exploration Rovers (MER) mission \citep{MERdataset}, from the Curiosity rover from Mars Science Laboratory (MSL) mission \citep{wagstaff2018deep}, \citep{MSLdataset} and from the HiRISE sensor on the Mars Reconnaissance Orbiter (MRO) \citep{HiRISE}. These datasets comprise images of landmarks on the Mars surface and have their own associated label, based on which landmark is represented. The label of the landmark is a crucial task in VBN, which exploits the recognition of specific features in the surrounding environment.\\
Differently, no relevant work of dataset generation or even AI application has been found, to the best of the author's knowledge, in the field of FDIR for space-employed camera sensors. Therefore, the present research contributes to an innovative FMEA of camera sensors, mainly leveraging industrial expertise and previous space exploration missions and the respective camera design documents, where available. Cassini \citep{west2010flight}, Hayabusa2 \citep{yamada2023inflight} and OSIRIS-REx \citep{bos2018touch}, \citep{bos2020flight} are examples of interplanetary exploration missions presenting similar traits to \textit{Astrone KI}. While Cassini does not include any landing, Hayabusa2 and OSIRIS-REx actually comprise a landing phase, but they can all contribute to the identification of failure modes which also apply to \textit{Astrone KI}. \citet{west2010flight}, \citet{yamada2023inflight}, \citet{bos2018touch}, \citet{bos2020flight} describe the design challenges of the camera sensors of the mentioned missions, reporting useful information for the occurring faults. Specifically, all the cameras focus on the analysis of the foreseen broken pixels rate, making considerations on the mission lifetime and the criticality of this fault for the navigation algorithm. Besides, strategies to reduce the occurrence of other fault cases are described. Inner and outer coating of the camera and the lenses, lens hoods and other mission-specific countermeasures can be found in the mentioned literature. Particularly useful is the work done in Cassini (Figure 34 in \citet{porco2004cassini}, Figure 18 in \citet{west2010flight}), where the analysis of reported images has highlighted typical fault patterns due to broken pixels and straylight.\\
Concerning other fault cases occurring in space cameras, the present research had to focus on other domains to retrieve applicable simulation strategies. \citet{chapman2011predicting}, \citet{chapman2013empirical}, \citet{chapman2018exploring}, \citet{chapman2022image}, \citet{chapman2023image} are a series of works about the occurrence of broken pixels in commercial cameras, that were extensively employed to retrieve a simulation methodology applicable to the space domain. \citet{king2001game}, \citet{maughan2001texture}, \citet{kilgard2000fast} are pioneering works in the field of straylight simulation, and were employed in the present work to guide the development, as the same methodologies are applicable to the space domain too. Moreover, the straylight field reports the presence of datasets in literature, addressing the flares removal task, targeting images from a variety of real-life scenarios \citep{wu2021train}, \citep{dai2022flare7k}, \citep{dai2023flare7k++}. These datasets can be efficiently used for pre-training or testing of specific AI solutions targeting flares removal and employing the dataset proposed in this work. 

Other faults presented in this work employ simulation strategies derived from the industrial expertise of the authors, as proper literature describing the fault occurrence and/or the simulation strategy is limited.

\section{Environment Description and Considered Fault Cases} \label{sec:envdesc_selfail}

\begin{figure}[h!]
    \centering
    \includegraphics[width=\linewidth]{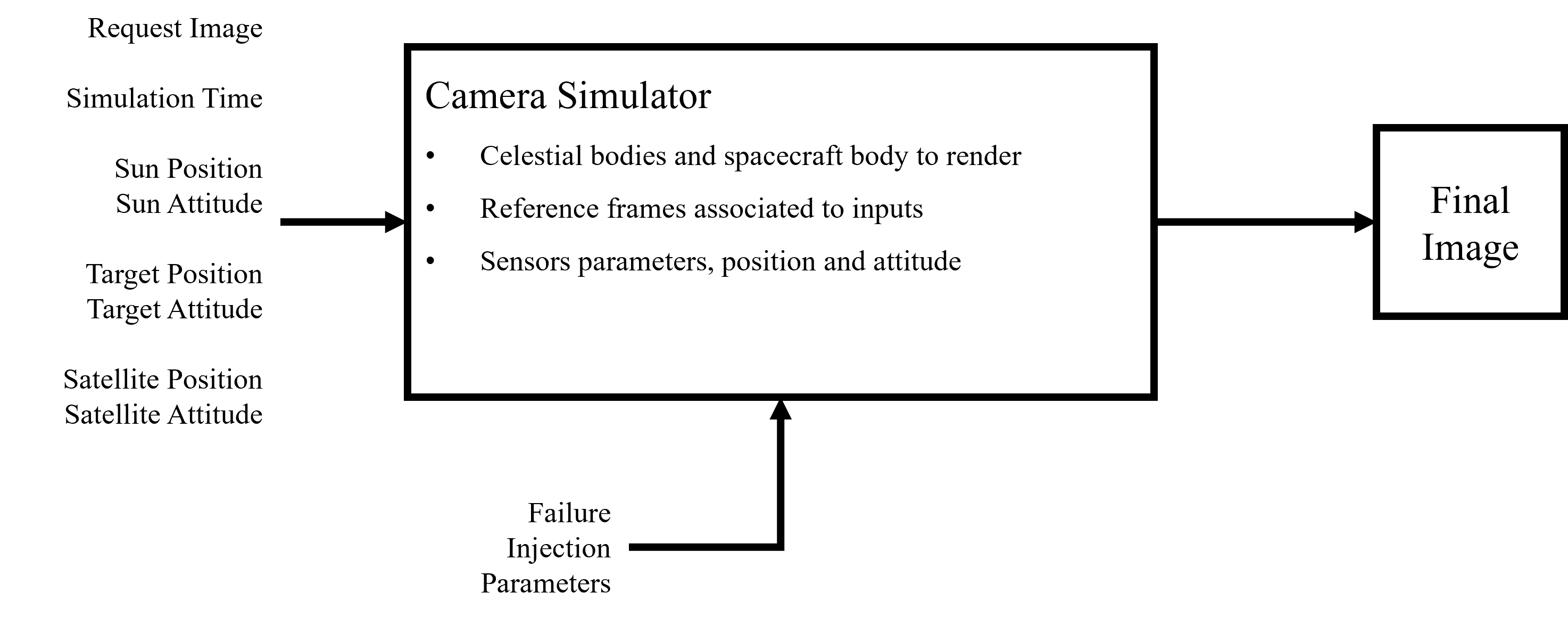}
    \caption{Schematic representation of the CamSim infrastructure in Simulink.}
    \label{fig:camsim_simulink}
\end{figure}

The present study focuses on the comet 67P/Churyumov-Gerasimenko, whose shape model is publicly provided by ESA based on the data collected by the Rosetta mission \citep{preusker2017global}, in the form of an object file (.obj). The comet is rendered in OpenGL via the Camera Simulator (CamSim) developed by ASTOS Solutions GmbH \citep{eggert2015imaging}, which enables the environment generation, as well as the simulation of the spacecraft and its sensors. The CamSim software also includes a Simulink interface, which is schematically shown in Figure \ref{fig:camsim_simulink}.

\begin{figure}[htb!]
    \centering
    \includegraphics[width=0.3\linewidth]{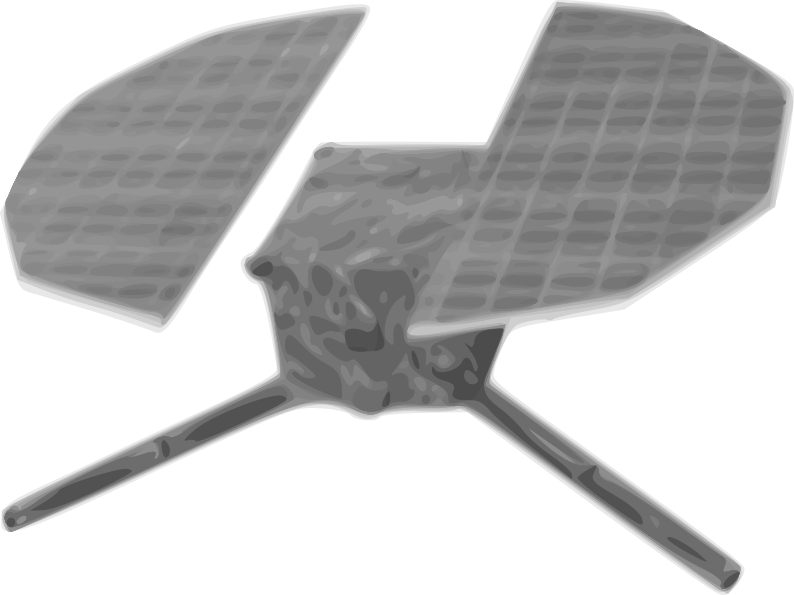}
    \caption{Simplified Astrone KI Model for Simulation in CamSim.}
    \label{fig:astrone_vehicle}
\end{figure}

In the case of \textit{Astrone KI}, Simulink is employed for the spacecraft simulation loop and to command the different CamSim parameters. These parameters define the celestial bodies and the spacecraft, with additional components eventually present (e.g. solar panels), to be rendered in the images. Celestial bodies simulation includes only the Sun, given the interplanetary focus of the present work. The Sun can be rendered without a specific object file, as it is already embedded in CamSim source code, and its position can be commanded along the simulation in a user-specified reference frame. Conversely, the comet (\textit{Target body} from now on) and the spacecraft rendering require  a custom-defined object file, including accessory components e.g. solar arrays and the landing gear, which can also be commanded in terms of position and attitude with respect to specified reference frames. A mock-up of the vehicle is shown in Figure \ref{fig:astrone_vehicle}.

The positions and attitudes employed to define the different simulation components can be entered in CamSim as defined in different reference frames. The Solar System Barycenter's (SSB) is implicitly defined in the simulator, while custom-defined frames can be related to the celestial or target bodies presence and modelling.\\
Finally, the definition of a Camera and/or LiDAR sensor is required to render the images of the surrounding environment. These sensors can be defined as attached to other bodies (e.g. spacecraft) and their orientation with respect to the respective Body-Fixed frames. Moreover, they can be setup based on pre-defined sensor-specific features.

On top of the realistic environment and sensors simulation, CamSim includes a variety of features to enable the simulation of camera faults that are widely employed in the present study. These features are:
\begin{itemize}
    \item Pixel Overwriting: the simulator allows to define textures of user-defined values that are merged to one or multiple color channels of the images.
    \item Gaussian Blurriness: a Gaussian filter can be applied to the rendered image to simulate a blurriness effect with a user-defined intensity.
    \item Straylight simulation: the simulator implements realistic rendering of user-defined textures representing solar flares and glares in the simulated images.
\end{itemize}

\subsection{Considered Fault Cases} \label{subs:cons_fail}
The following section presents a series of faults arising from the present research, that were deemed significant to investigate since they affect Vision-based Navigation systems. The approach to the faults identification started by leveraging the industrial expertise of visual sensors engineers to identify common failure modes, integrating it with an extensive literature research. Successively, the effects of the identified failure modes on the VBN pipeline are considered. The present work does not make a distinction among critical and non-critical faults, but rather it aims to shed light on the variety of faults that can occur in camera sensors regardless of their criticality. However, the faults are tailored on the \textit{Astrone KI} use-case, meaning that they take into account the specific mission scenario. Other interplanetary exploration missions or even targeting different scenarios (i.e. planetary exploration, Earth Observation) may need additional or reduced failure modes consideration. The main drivers derived from \textit{Astrone KI} are mainly the interplanetary unshielded radiation environment, the prolonged mission lifetime, the fast dynamics and the expected latency of performing onboard procedures commanded from the ground.

In conclusion, the description of each fault presented hereafter is structured as follows. An overview of the fault effects at system level is given, followed by a detailed description of the causes and effects at sensor level, which constitute the main driver to the simulation strategy described in Section \ref{subs:fail_inj}. Finally, common mitigation strategies are described, with references to existing missions making use of them.

\subsubsection*{Dust on Optics}
    The problem of dust on optics is particularly critical for optical systems, because it represents a consistent reduction of the visibility. It is especially present in landing vehicles, where dust may raise from the ground in the landing site. The dust rises by effect of the force exerted by the thrusters and electrostatically sticks to the optical lenses, accumulating in grains. The grains can aggregate in different quantities and acquire different shapes, determining a more or less consistent shadowing of the lens area. In \textit{Astrone KI}, differently from other existing exploration missions, the requirement of performing different relocations over the comet surface make the dust on optics significantly critical, because it must be always possible to trust the output of the optics in sight of the next use.
    
    There is no standardized way to avoid deposition of dust on lenses in space, since missions exploiting vision-based systems for landing are limited and not standardized, thus each one presents its specific solution. The closest mission concept to \textit{Astrone KI}, which performs autonomous relocation over an extra-terrestrial surface, is the Mars helicopter \textit{Ingenuity}. It is a technology demonstrator to prove the capability of flying over the Mars surface employing Vision-based Navigation algorithms \citep{bayard2019vision}. The vehicle deals with the problem of dust on optics in its landing and take-off phase, relying only on its Inertial Measurement Unit (IMU) to guide the trajectory, as the camera would not acquire meaningful data due to the rising dust. %Besides, it utilizes a car-like wiper to clean out the deposited dust on the camera lens after each landing (\textcolor{red}{can't find a reference})
    
\subsubsection*{Broken Pixels}
    Broken pixels are a diffused problem in camera sensors, equally affecting commercial and space-grade cameras. The defects can assume different patterns in the image, based on their root cause, with the result of limiting the camera visibility and the capability to distinguish details. Besides, infinite and/or zero value in the pixel output may also be misprocessed in the navigation algorithm, leading to incorrect results or ultimately processing faults. 
    
\begin{figure}[h!]
        \centering
        \includegraphics[width=0.6\linewidth]{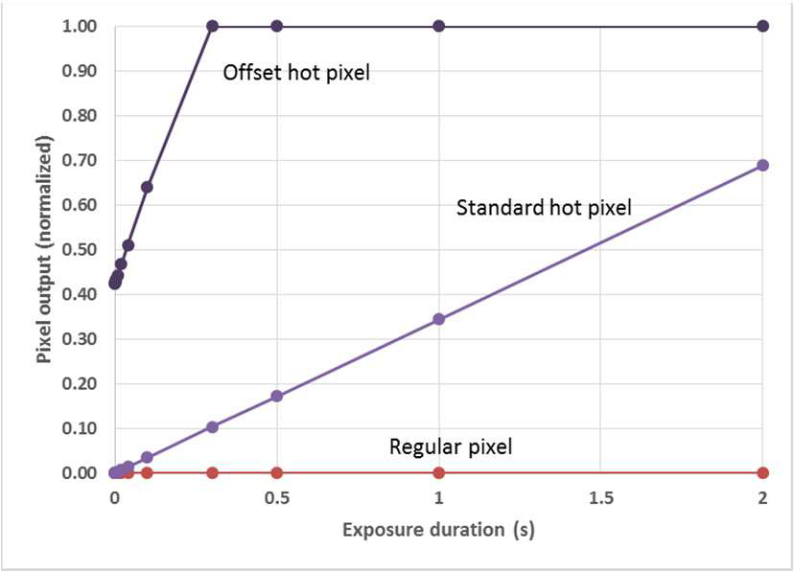}
        \caption{Comparison of Regular and Hot Pixels Sample Dark Response. The regular pixel keeps its output constant over the Exposure Time, while hot pixels present a linear behaviour\citep{chapman2022image}.}
        \label{fig:bp_response}
    \end{figure}

    A first differentiation includes defects caused by the external environment or internal electronic components. The external environment affects pixel defects by means of the space radiation impinging the detector surface. Depending on the specific mission profile, e.g. LEO, GEO or interplanetary orbits can experience significantly different radiation environments \citep{Jun2024}, determining the number of defective events. Particles impinging a pixel cause a modification in its response to light. Pixels whose properties have been modified by a transient radiation particle are called \textit{Hot Pixels}. In Figure \ref{fig:bp_response}, it is possible to observe different responses of hot pixels as measured in a fully dark environment (\textit{dark response}).
    
    Increasing the Exposure Time, the pixel output will increase and potentially saturate if the impinging radiation has introduced an offset in its response curve. Besides, the two extreme cases of a fully hot and fully dead pixels, called \textit{Salt\&Pepper Defect}, can be present, but they have been observed to be much less frequent than what has been measured because they are often confused with offset hot pixels \citep{leung2007quantitative}. This last statement is derived from the works of \citet{leung2007quantitative}, \citet{leung2009statistical}, \citet{chapman2011predicting}, \citet{chapman2013empirical}, which proved it in smartphone commercial cameras, but has no further evidence in space, where the radiation effect is eventually stronger. This latter feature may lead to increased \textit{Salt\&Pepper Defect}) rate, but this has not been verified yet. Further studies \citep{chapman2022image}, \citep{chapman2023image} also suggest a correlation between the broken pixel occurrence, empirically showing that the occurrence of a couple of broken pixels in a 3x3 neighborhood is higher than the probability expected only considering the generalised birthday problem of random processes.
    
\begin{figure}[h!]
        \centering
        \includegraphics[width=0.6\linewidth]{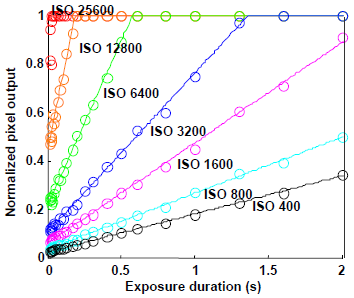}
        \caption{Dark Response of the Same Hot Pixel at Varying ISO \citep{chapman2013empirical}.}
        \label{fig:pixelresponse_ISO}
    \end{figure}

    Camera features affecting the broken pixels occurrence include Exposure Time and sensor sensitivity to light (ISO), as these two variables control the amount of light absorbed by the detector. This, in turn, determines the characteristics of the hot pixel in the final image. Based on the response curve in Figure \ref{fig:bp_response}, the response of hot pixels increases with the Exposure Time, eventually leading to increased brightness. Figure \ref{fig:pixelresponse_ISO} shows the hot pixels correlation with ISO, which determines a generic increase of the pixel output with increasing ISO.

    Other pixel defects occurring in camera sensors are related to random faults in the electronics, especially the readout components. \textit{Readout} refers to the process of retrieving the current measure from the single pixels to process the final image. There are different readout mechanisms based on the existing sensors architectures \citep{waltham2013ccd}. In space, Charge-coupled Device (CCD) sensors were preferred to Complementary Metal-oxide Semiconductor (CMOS) sensors \citep{bos2018touch}, \citep{bos2020flight} in interplanetary missions for their inherently higher sensitivity \citep{waltham2013ccd}, \citep{yamada2023inflight}, \citep{west2010flight}, while more recent missions shifted to CMOS detectors which in turn provide lower cost and power consumption \citep{waltham2013ccd}. In addition, the two kinds of detectors present different readout mechanisms, which in turn affect the associated failure modes. In CCD sensors charges shift towards the readout component and are read in series, while in CMOS they are read in parallel per row (or column). More details on this two construction choices are out of the scope of this work and can be found in \citet{waltham2013ccd}.

    An common effect in cameras employing both CCD and CMOS sensors is the spreading of a defect to other pixels than the broken one, which is caused by \textit{demosaicing} algorithms. \textit{Demosaicing} refers to algorithms employed in the camera software responsible of postprocessing the output of the detector, in order to retrieve the color value of a single pixel from its current output. As the detector is made by different color filters, responsible of grasping different light colors, the output of the different filters is interpolated to retrieve a final color for the associated pixel \citep{chapman2022image}. This interpolation involves neighbouring pixels in a way that whenever a pixel is broken, its defective output will corrupt also the pixels around it.  Different algorithms of demosaicing exist, where \textit{bilinear interpolation} is the simplest \citep{longere2002perceptual}. \citet{chapman2023image} provides an example of how bilinear interpolation spreads the defect of a single pixel in its von Neumann neighbourhood (i.e. top, bottom, right, left pixels).
    
    More defects in common between CCD and CMOS detectors are associated to other components taking part in the readout. Registers, multiplexers and row select transistors \citep{waltham2013ccd} regulate the row/columns to read out, selecting the pixel coordinates and shifting them along the process. A defect in these components can generate full row or column defects, which means completely saturated or dead lines of pixels crossing the whole image. These defects can occur during flight operations after an unforeseen fault in the electronics. By construction, Full-frame and Frame-transfer CCD sensors \citep{waltham2013ccd} can experience mutually exclusive row or column defects, while Interline-transfer CCD and CMOS \citep{waltham2013ccd} can be affected by both row and columns at the same time.

    The problem of defect pixels, and in general of increasing dark currents in CCD and CMOS detectors, is faced via calibration procedures that strongly depends on the different mission where they are applied. \citet{west2010flight}, \citet{yamada2023inflight}, \citet{bos2018touch}, \citet{bos2020flight} give an overview of how the calibration of the camera sensors of the relative missions is carried out. Data about the number of broken pixels occurring in OSIRIS-REx and Cassini \citep{bos2018touch}, \citep{bos2020flight}, \citep{west2010flight} are also provided. Both missions report the presence of many broken pixels and try to correct them via in-flight calibration, though representing no critical problem in the respective use. In \citet{bos2020flight} it is observed that a concentration of $>3$ broken pixels in a close neighborhood can cause misdetection of objects on the asteroid surface, a condition that applies also to the \textit{Astrone KI} case and deserves investigation. 
    
    In general, it is possible to retrieve useful information from the mentioned missions to analyze the \textit{Astrone KI} defect pixels fault, but significant differences apply. The main point of divergence is that the effects of the defect pixels cannot be clearly predicted in \textit{Astrone KI}, as the mission is currently in Phase A with an operational lifetime still to be defined. Moreover, the insufficient number of similar missions which may act as a baseline, prevents an estimation of this parameter. As a consequence, variables such as the lifetime of the camera sensor and the design of the communication link with ground stations are also not clearly defined, while they strongly affect the evolution of the defect pixels fault. Indeed, faults occurrence increases with a greater radiation dose absorbed by the detector, while it decreases if in-flight calibration procedures can be performed.
        
\subsubsection*{Straylight}
\begin{figure}[h!]
        \centering
        \includegraphics[width=0.6\linewidth]{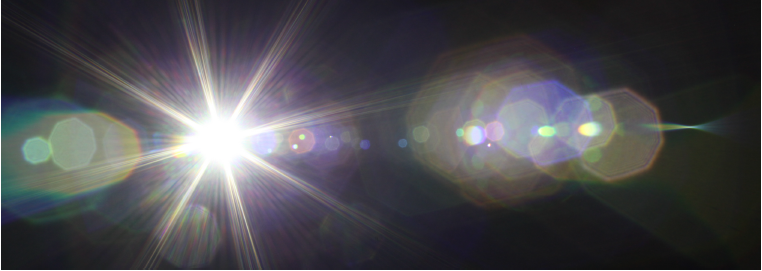}
        \caption{Example of Physically Rendered Flares in \citet{hullin2011physically}.}
        \label{fig:phys_flares}
    \end{figure}

    \textit{Straylight} is a generic term referring to peculiar artifacts in camera images generated by unintended reflections within the camera lenses of the light coming from objects situated inside or slightly outside the Field of View (FoV). The straylight artifacts include \textit{flares} and \textit{glares}. Flares are sudden burst of light which realize in discrete textures originating from the light source. Glares are characterized by a diffuse spreading of the light. The causing factors of these flares and glares are tightly bound to the specific instrument model and to the lenses arrangement. In space, straylight can be generated by reflection of any source of light that can appear in the FoV, namely the Sun, celestial bodies and stars. Its main consequence is the reduction of the capability of the camera to clearly distinguish objects, covering them with artifacts of various brightness and colors. The effects of straylight on camera lenses have been studied in literature across different fields such as cinema, animation and video games. First approaches in straylight simulation employed fixed textures blending with the image \citep{king2001game}, \citep{maughan2001texture}, \citep{kilgard2000fast}. Textures were generated \textit{a priori} with respect to the images and could assume shapes of e.g. \textit{rings}, \textit{halos} and \textit{orbs}. They were positioned in the image on a line linking the light source to the center of the image, while their brightness varied according to the distance from the light source. More recently, \citet{keshmirian2008physically}, \citet{hullin2011physically}, \citet{lee2013practical} proposed refined methodologies based on physical lens simulation, marking the current standard in the state of the art (Figure \ref{fig:phys_flares}).
    
    The effect of straylight in space camera sensors can be generally attenuated by using barrel designs, anti-reflective coatings and lens hoods. Cassini reported the presence of straylight during the in-flight calibration procedure of its science subsystem cameras \citep{west2010flight}, having no protection device installed. In Hayabusa2 the problem is mitigated by installing L-shaped baffles and an inner coating to the camera housing \citep{jaumann2017camera} in its Mobile Asteroid Surface Scout (MASCOT) camera (MasCam). Nevertheless, the listed attenuation measures, while reducing the straylight phenomenon, cannot eliminate it completely due to its inherent bond with the light source presence. In \textit{Astrone KI}, the potentially fast vehicle dynamics relying on VBN algorithms makes straylight a significant issue. Covering objects by reflexes and reducing the contrast of the whole scene are primary aspects of this fault case that may lead to navigation misfunction and fault. Therefore, an accurate detection of straylight is necessary, eventually followed by its removal as recovery. Ultimately, operational constraints can be established to reduce the occurrence of flares and glares, e.g. limiting the range of possible trajectories.
    
\subsubsection*{Vignetting}    
    The vignetting effect is a well-known inherent effect of any existing camera system, and it is caused by the light hitting the sensor at different angles, depending in turn on the specific lenses allocation. Lower angles are associated to off-axis sources and longer distances for the light to cover to impinge the detector, determining a reduction of the transferred power. As a consequence, Vignetting causes a generalized darkening of the edges of the image, which can be described by the \textit{cosine-fourth law} \citep{kerr2007derivation}. This law states that the illuminance degradation of an off-axis object ($E_{\Theta}$) with respect to an on-axis object ($E_{0}$) with the same area can be computed as: 
    \begin{equation} \label{eq:cos4_law}
        \centering
        \dfrac{E_{\Theta}}{E_{0}}=\cos^4{\Theta}    
    \end{equation}
 
    where $\Theta$ is the off-axis angle. The \textit{cosine-fourth} law is valid for \textit{natural vignetting} only, which is present in any camera sensor. Other vignetting effects include \textit{mechanical vignetting} and \textit{optical vignetting}. The former is related to mechanical obstruction of the lens (e.g. due to wrongly designed lens cases) and can easily be avoided in the manufacturing phase. The latter is related to the specific configuration of the lenses, but provides the same edges-darkening effect of the natural vignetting. For the sake of the present work, optical vignetting is not covered in detail, but since it provides an edge-darkening effect as well as natural vignetting, the \textit{cosine-fourth} law is assumed to describe both effects.
    
    As with straylight, avoiding the occurrence of Vignetting is fundamentally infeasible, given its intrinsic link to the presence of a light source. However, post-processing methods exist to mitigate this effect, typically through software corrections during image acquisition. These approaches identify Vignetting and adjust the brightness gradient accordingly. Utilizing AI for Vignetting detection could facilitate faster and more effective correction, making it valuable to include this effect in the present dataset.

\subsubsection*{Optics Degradation}
    Optics degradation causes a gradual worsening of the optical performances of a system of lenses along the operational time. This phenomenon is usually associated to environmental factors such as atomic oxygen, UV radiation, outgassing and cross-contamination, charged particles, impact of micrometeoroids and/or space debris and dust deposition. These effects are very common in the space environment and their effects are usually estimated in the mission design phase to compensate or shield the affected components. The major issue for optical sensors is the changes in the surface properties of the external lenses, which affect the way light is transferred across them. \citet{garoli2020mirrors} provide a comprehensive review of factors affecting mirrors for space telescopes, as the same criticalities apply to camera sensors, identifying the most common degradation effects. As these effects can be diverse on final images, in this work a diffused image blurriness is considered as only consequence. A more detailed study of degradation effects can be a promising field of future work.
    
    The implications of a blurred image include difficult object recognition and a general details distinction, causing the navigation algorithm to not function properly. Specifically to VBN solutions, blurred landmarks can lead to imprecise or failed pose estimation and edge detection, requiring the presence of a fallback algorithm and/or an efficient FDIR strategy. The usual way of contrasting optics degradation in space is to adopt countermeasures in design phase. Typically, lens coatings, lenses materials selection and surface treatments are designed to resist the aforementioned degradation causes, employing prediction models to determine the effects and improve the components lifetime.
    %misc: $-$

\section{Dataset Generation} \label{sec:datagen}
\begin{figure}[h!]
    \centering
    \includegraphics[width=\linewidth]{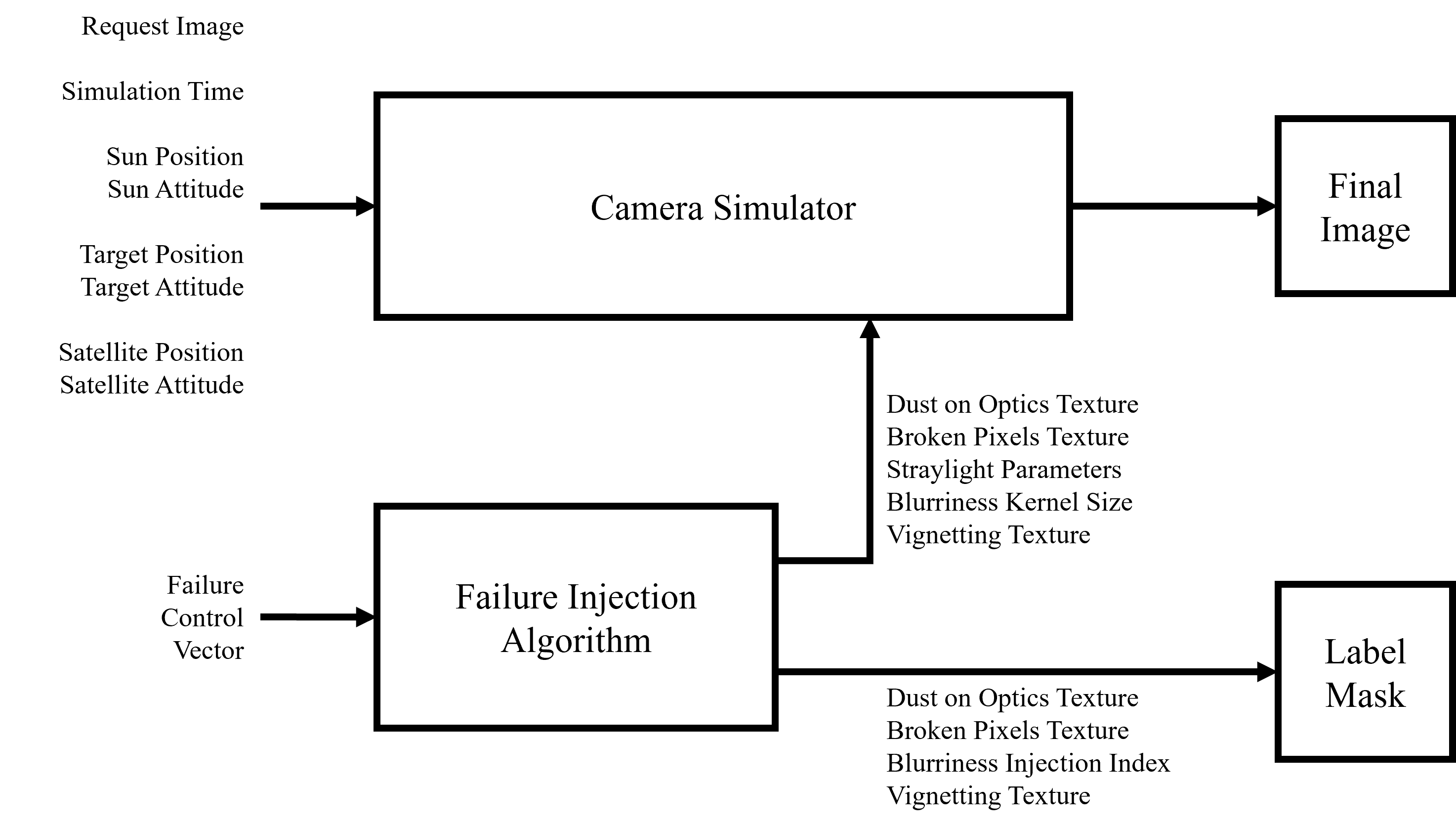}
    \caption{Overview of the Dataset Generation Environment.}
    \label{fig:simulink}
\end{figure}

The simulation infrastructure for the dataset generation is depicted in Figure \ref{fig:simulink}, where the CamSim block for the image generation is located in the upper part, while the lower part is dedicated to the fault injection and simulation mechanism. The lower part is itself divided into a first stage where the injection of the faults is commanded, a second stage where the faults parameters are computed, and finally a third stage dedicated to the mask generation.

The environment simulation in CamSim includes the Sun, the comet (i.e. target body) and the spacecraft body, with their respective \textit{.obj} files. The position of the Target is given by its ephemerides at a specified Julian date. The Target body rendering is divided in patches starting from the original full model \citep{preusker2017global} and the spacecraft motion is defined within the patch. 

\begin{figure}[h!]
    \centering
    {
        \includegraphics[width=0.45\linewidth]{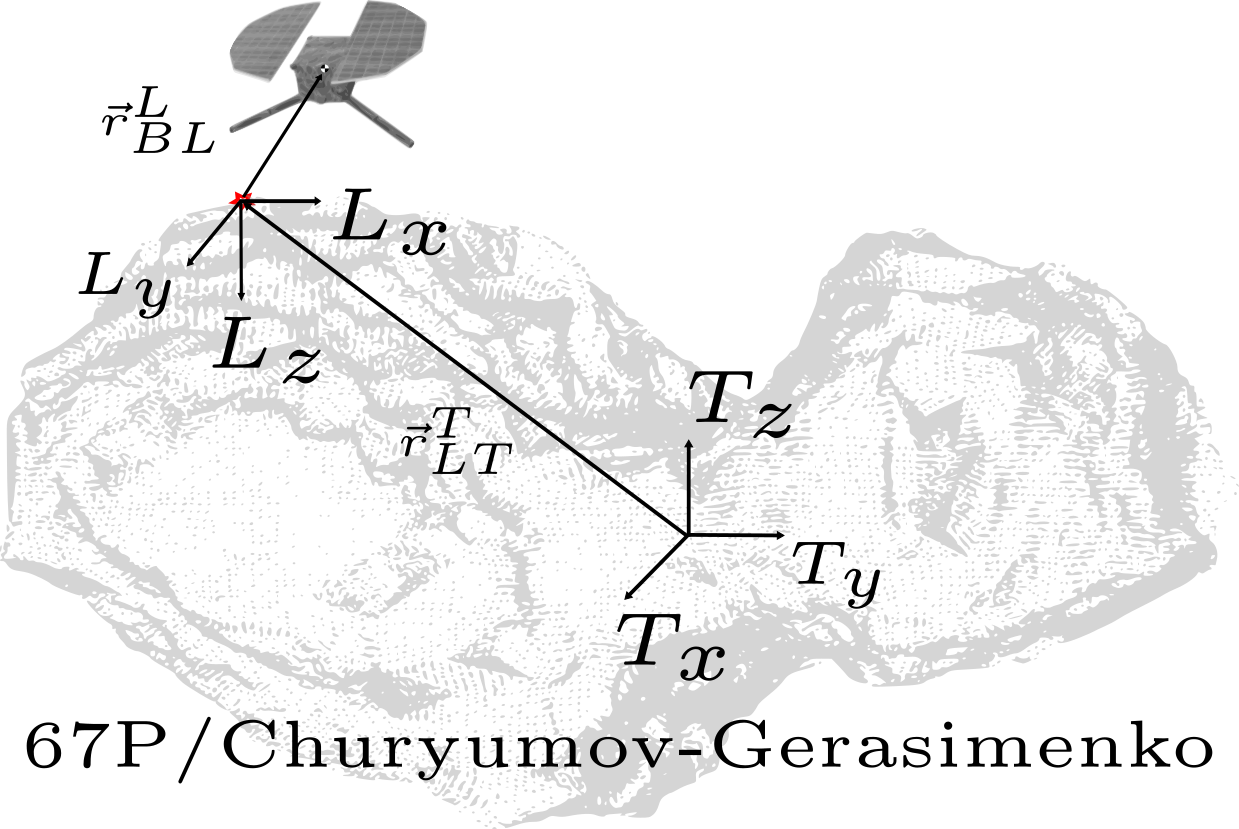}
    }
    \hfill
    {
        \includegraphics[width=0.45\linewidth]{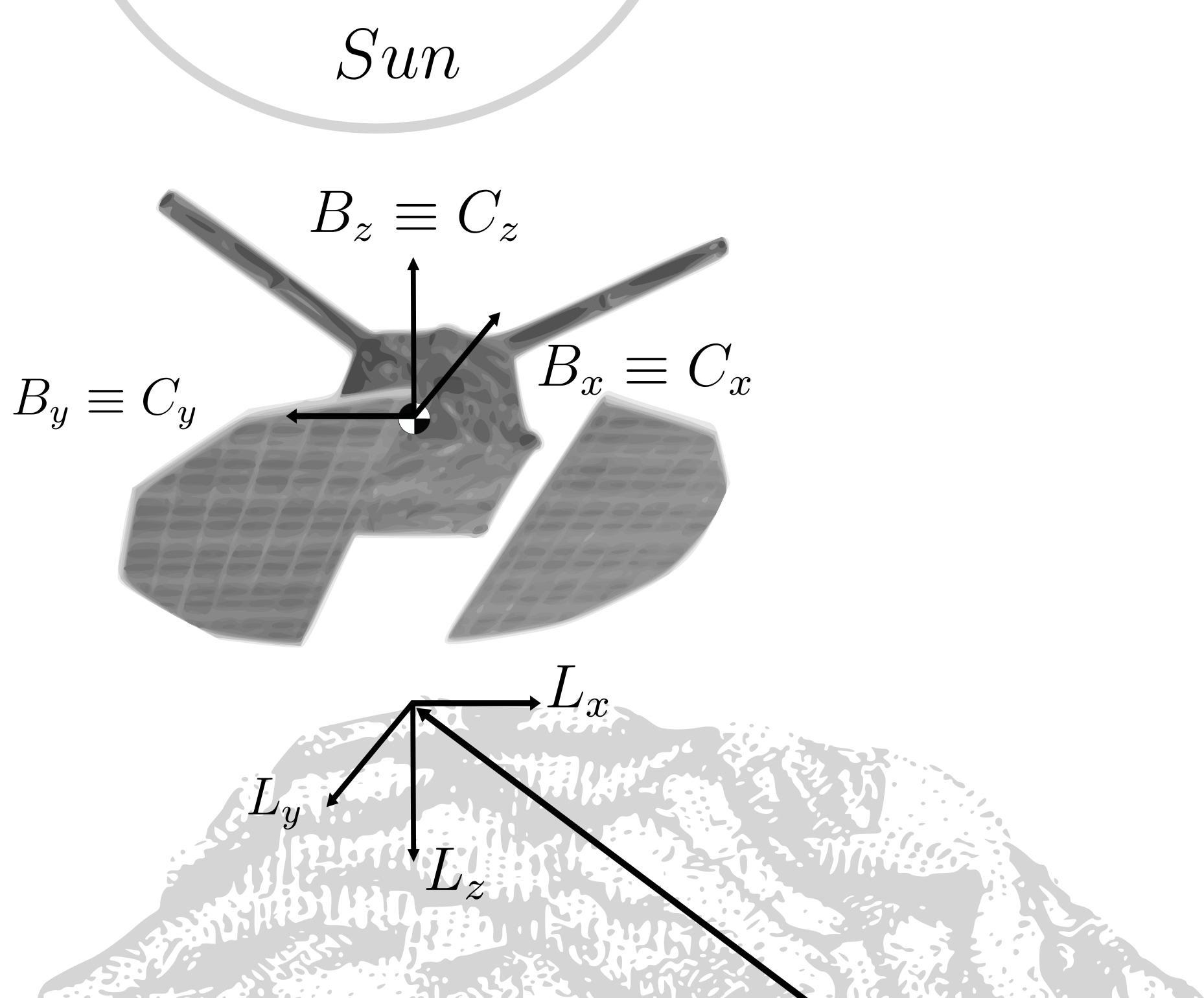}
    }
    \caption{Reference Frames Employed in the Dataset Generation, Adapted from  \citet{olucak2023sensor}.}
    \label{fig:TLB_frames}
\end{figure}

The reference frames introduced to define the motion of the rendered bodies are presented in detail in \ref{app:A}. More details are needed to describe the Local-Level frame with respect to the others, as it plays a crucial role in the definition of dataset variables which will be employed in the simulation. The Local-Level frame is computed locally on the terrain patch, via a separated procedure that slices the patch from the full comet model and interpolates the model points to enhance the resolution. The Local-Level frame coordinates in each point of the enhanced-resolution map are then computed based on the gravity vector. The gravity data are provided together with the global comet model in \citet{preusker2017global}. The details of the patching procedure are out of the scope of this paper. \\
Figure \ref{fig:TLB_frames} gives a graphical overview of all the reference frames involved in the simulation. Specifically, $\mathcal{T}$ indicates the Target-Body Centered Target-Body Fixed Frame, $\mathcal{L}$ is the Local-Level Frame, $\mathcal{B}$ and $\mathcal{C}$ are the Body-Fixed and Camera-Fixed Frames respectively.

\subsection{Image Acquisition} \label{subs:strat_sampl}

The image acquisition procedure outputs images by setting randomized parameters for the environment, sensor and fault variables. No specific spacecraft trajectory is assumed as input, leading subsequent images to be uncorrelated. This assumption avoids on one side the presence of biases in the data derived from the constrained variables values. On the other side, it limits the AI application employing the dataset to process only one image at a time. 

First, representative independent variables of the environment and  parameters of the camera sensor shall be identified. The definition of the camera sensor in \textit{Astrone KI} is not detailed up to the identification of a specific camera model or set of lenses, hence none of them are employed in the dataset generation. The definition of the optical parameters is kept as generic as possible, aiming to embed in the data the representation of diverse hardware setups, and boost the applicability range of the trained AI.

The camera and the image acquisition parameters employed in the present work are specified in Table \ref{tab:camera_par}.
\begin{table}[h!]
    \centering
    \begin{tabular}{c|c}
        \textbf{Parameter} & \textbf{Value} \\
        \hline
        FoV                 & 65 deg \\
        Resolution          & $1024\times1024$ pixels \\
        Color Depth         & 8 bits per channel \\
        Number of Channels  & 4 (R, G, B, I) \\
        Values per Channel  & 256 ($0-255$)\\
        Image Format        & RGBI \\
        Color Range         & RGB \\
    \end{tabular}
    \caption{Camera and Image Parameters.}
    \label{tab:camera_par}
\end{table}

The environment variables that influence the scenario and must be varied to generate a comprehensive dataset include the relative Sun-comet position, the satellite position on the surface of the comet and the satellite attitude. These variables are deemed sufficient to express a representative dataset. Due to the presence of the straylight fault (Section \ref{subs:cons_fail}), which is bounded to the Sun position, a classic random sampling over the unconstrained boundaries of variation of the dataset variables would not be an efficient strategy to apply. Specifically, this fault occurs when the Sun stays in the camera FoV or in its proximity. Hence, utilizing a classic random sampling would inject an unpredictable number of straylight features in the dataset, corresponding to those image where the Sun randomly occurs in a suitable position. This behaviour would result in an unpredictable balance of the classes in the dataset, ultimately leading to an insufficiently representative number of certain faults with respect to others. Conversely, the other faults listed in Section \ref{subs:cons_fail} do not depend on the Sun position, thus can be randomly injected across all the generated images, as detailed in Section \ref{subs:fail_inj}.

\begin{figure}[h!]
    \centering
    \includegraphics[width=\linewidth]{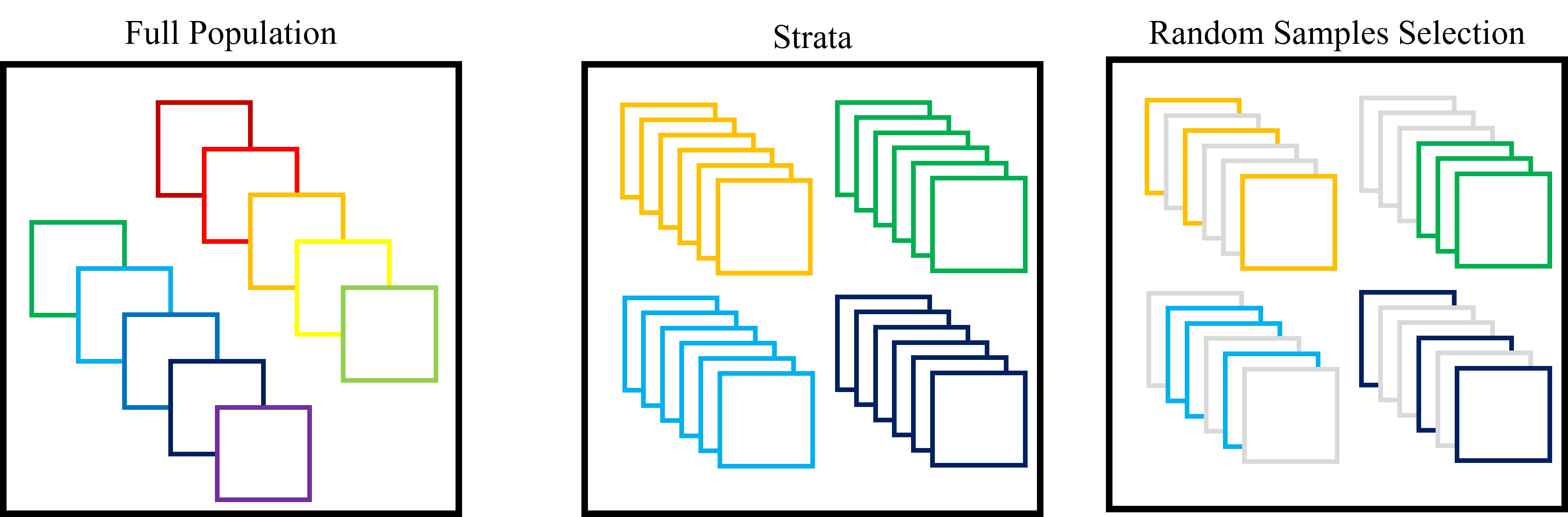}
    \caption{Stratified Sampling Technique}
    \label{fig:strat_sampl}
\end{figure}

Stratified sampling (Figure \ref{fig:strat_sampl}) has been utilized to ensure uniformity across the fault classes. This methodology involves a two-stage sampling process. In the first stage, the dataset variables are constrained to include instances where the Sun is within specific boundaries of the Camera-Fixed frame. Random sampling is then conducted to capture images at various Sun positions, ensuring the presence of straylight. In the subsequent stage, the camera is directed away from the Sun, and random sampling is again employed to collect the remaining images for the dataset. The number of images acquired in both phases ensures an appropriate percentage of samples belonging to the straylight class.

The simulation setup requires the Sun position and the Target position and attitude, together with the spacecraft attitude, to be defined in the International Celestial Frame ($\mathcal{J}$), while the spacecraft position is defined in the $\mathcal{T}$ frame. The camera sensor is instead attached to the spacecraft, so its orientation is relative to the $\mathcal{B}$ frame. Although the initial Sun state is imposed in the $\mathcal{J}$ frame, its position with respect to the Target body is commanded in the $\mathcal{L}$ frame, employing Azimuth ($\rho$) and Elevation ($\gamma$). The major advantage offered by this reference frame resides in the intuitiveness, allowing to achieve a comprehensive coverage of the illumination conditions in the dataset. Remarkably, for reasons related to the CamSim implementation, the Azimuth and Elevation of the Sun position are translated into the relative Target body attitude, which is then employed in the simulator to obtain the defined illumination. The Sun position and the Target attitude are indeed coupled variables, thus can be used equivalently to the dataset generation purpose. The satellite attitude is set to rotate with respect to the Body-Fixed frame according to the Euler Angles $\psi$ (roll), $\theta$ (pitch) and $\varphi$ (yaw). The Camera-Fixed frame and the Body-Fixed frame are set as coincident as to the purpose of the dataset generation the specific allocation of the camera in the vehicle is irrelevant. The satellite position on the comet surface is varied in the Local-Level frame according to the definition of the single patches.

Due to the sampling methodology detailed earlier, it is convenient to start the image generation process by aligning the $\mathcal{L}$, $\mathcal{B}$ and $\mathcal{C}$ frames with their \textit{z}-axes pointing at the Sun, which means fixing the Sun in the center of the Camera FoV. This scheme allows a convenient definition of the boundaries of the dataset variables, especially the spacecraft attitude angles, where the Sun pointing corresponds to 0 deg.

The complete dataset generation procedure is detailed hereafter:
\begin{enumerate}
    \item Generate a random Target attitude to fix the illumination conditions.
    \item Retrieve the attitude of the spacecraft imposing the alignment of the camera with the Sun direction.
    \item Randomize the spacecraft attitude to keep the Sun in proximity of your camera FoV $\rightarrow$ Acquire image.
    \item Randomize the spacecraft attitude to keep the Sun outside your camera FoV  $\rightarrow$ Acquire image.
\end{enumerate}

Step 3 and 4 of the procedure differentiate by the boundaries of the variables, set to obtain Sun-facing or Sun-averted orientations. All the possible illumination conditions and attitude angles of the spacecraft are efficiently covered, ensuring sufficient variety of scenarios framed by the camera.

Table \ref{tab:bound_var} summarizes the dataset variables and their boundaries in the two stages of sampling.
\begin{table}[h!]
    \centering
    \begin{tabularx}{\linewidth}{X|X|X|X}
                                        &\textit{Sun-facing} Case   & \textit{Sun-averted} Case   &
                                        Reference Frame\\
         \hline
        Sun Azimuth $\rho$              & $0<\rho<2\pi$        & $0<\rho<2\pi$          & Local-Level, $\mathcal{L}$\\
        Sun Elevation $\gamma$          & $-\dfrac{\pi}{6}<\rho<\dfrac{\pi}{6}$  %$\pm \left( \dfrac{\pi}{2}-\dfrac{\pi}{6}<\rho<\dfrac{\pi}{2}+\dfrac{\pi}{6} \right)$ 
        & $-\dfrac{\pi}{6}<\rho<\dfrac{\pi}{2}$ & Local-Level, $\mathcal{L}$\\ %$-\dfrac{\pi}{2}-\dfrac{\pi}{6}<\rho<\dfrac{\pi}{2}+\dfrac{\pi}{6}$ \\
        & &   \\
        Spacecraft Roll Angle $\psi$     & $-FoV/2<\psi<FoV/2$       &  $FoV/2<\psi<2\pi-FoV/2$  & Body-Fixed, $\mathcal{B}$ \\
        Spacecraft Pitch Angle $\theta$  & $-FoV/2<\psi<FoV/2$       &  $FoV/2<\psi<2\pi-FoV/2$  & Body-Fixed, $\mathcal{B}$ \\
        Spacecraft Yaw Angle $\varphi$   & $0<\varphi<2\pi$      & $0<\varphi<2\pi$      & Body-Fixed, $\mathcal{B}$ \\
        & &  \\
        Spacecraft Position, $x_{B}$    & $-150<x<150 \quad[-]$ &  $-150<x<150 \quad[-]$ & Local-Level, $\mathcal{L}$ \\
        Spacecraft Position, $y_{B}$    & $-150<x<150 \quad[-]$ &  $-150<x<150 \quad[-]$ & Local-Level, $\mathcal{L}$ \\
        Spacecraft Position, $z_{B}$    & $0<x<100 \quad[-]$    &  $0<x<100 \quad[-]$ & Local-Level, $\mathcal{L}$ \\
    \end{tabularx}
    \caption{Boundaries of Dataset Variables.}
    \label{tab:bound_var}
\end{table}

The Sun elevation in the \textit{Sun-facing} phase presents narrow bounds and implies that the Sun always stays close to the horizon of the landscape seen by the \textit{Astrone KI} camera. This choice aims to obtain a balanced percentage of images depicting only the sky and images where the terrain is also visible. A wider choice of the $\gamma$ interval would lead to increased percentage of sky-only images, penalizing to the ones where the terrain is present. The choice of the specific values has been done empirically as there is no need for an accurate computation, as long as both sky-only and terrain images present statistically significant samples. In fact, both cases are significant from an anomaly detection point of view, especially sky-only images which cover those fault cases causing loss of nominal attitude and off-axis camera pointing.

\subsection{Faults Injection} \label{subs:fail_inj}
% Explain the fault injection mechanism.\\
% Start explaining the fault control vector and which constraints on the faults apply (e.g. broken pixels due to different readout mechanisms can't go together).
% Explain how faults are randomized and how boundaries to the parameters introduced above are derived.
The mechanism of faults injection in the whole dataset aims to obtain a balanced dataset including sufficiently representative image samples per fault. This requirement is key to make sure that all the faults will be correctly learnt by an AI employing the dataset. However, further balancing depends on the task that the algorithm accomplishes, and is not in the scope of this work.

The present dataset generation method allows to set the percentage of faulty over nominal image samples, where the faulty samples contain uniformly distributed fault cases. This approach enables sufficient freedom to be used in many different tasks. For instance, setting a percentage to 0.5 yields a balanced dataset for binary (i.e. faulty, not faulty) classification, while a percentage of \textit{1/(\# fault classes)} can be more suitable to multi-class classification.

The fault injection is regulated by a matrix defined once at the beginning of the whole generation process, containing flags for commanding the occurrence of the single faults in each of the generated image samples. This matrix has size \textit{\# fault classes} $\times$ \textit{number of samples} and is generated based on these two parameters. The matrix generation does not include the straylight fault, which is appended as last row and it is constructed as shown in Equation \ref{eq:straylight_inj}:
\begin{equation} \label{eq:straylight_inj}
    \begin{cases}
        1, & \text{for } i = 1 : n_\text{straylight} \\
        0, & \text{for } i = n_\text{straylight}+1 : n_\text{tot}
    \end{cases}
\end{equation}
where $n_{tot}$ is the total number of samples to generate, while $n_{straylight}$ is the number of samples commanded to face the Sun and derives from the specified nominal-to-faulty samples percentage. In the resulting dataset, the first $n_{straylight}$ images include a straylight fault, while the other faults are randomly distributed. This specific construction descends from the stratified sampling technique (Section \ref{subs:strat_sampl}), which requires the straylight fault to be simulated first in the generation procedure. The methodology exposed for the straylight fault is generalizable to any other fault related to the Sun presence.

\subsection{Faults Simulation}
This section introduces the simulation strategy of the faults and the relative parameters. The approach makes heavy use of the capability of CamSim to overwrite pixels value, allowing to define textures that can afterwards be rendered onto the image. This feature is useful to simulate Dust on Optics, Broken Pixels and Vignetting, while Straylight and Optics Degradation rely on their specific features of the simulator.

    \subsubsection*{Dust on Optics}
\begin{figure}[h!]
        \centering
        \begin{subfigure}[b]{0.23\textwidth}
            \centering
            \includegraphics[width=\textwidth]{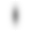}
        \end{subfigure}
        \hfill
        \begin{subfigure}[b]{0.23\textwidth}
            \centering
            \includegraphics[width=\textwidth]{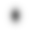}
        \end{subfigure}
        \hfill
        \begin{subfigure}[b]{0.23\textwidth}
            \centering
            \includegraphics[width=\textwidth]{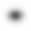}
        \end{subfigure}
        \hfill
        \begin{subfigure}[b]{0.23\textwidth}
            \centering
            \includegraphics[width=\textwidth]{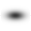}
        \end{subfigure}
        \caption{Sample Textures of Dust Grains with the respective parameters. From left to right: \textit{Max of the Gaussian Distribution} (100, 133, 166, 200); $\sigma_{xx}$ (3, 4, 5, 6); $\sigma_{yy}$ (3, 4, 5, 6); $\sigma_{xy}$ (0, 0, 0, 0)}
        \label{fig:dust_grains}
    \end{figure}

    % Each dust grain is modelled as a Gaussian texture to be overlapped to the blurriness challenge. Overlapping occurs via mixture blending. The texture shape and intensity can be varied acting on the mean and $\sigma$ of the Gaussian surface.
    The modelling of the Dust on Optics consists in the creation of a texture that reproduces the typical artifacts of dust grains depositing on the lens surface. The texture is overlapped to the Intensity channel of the single image and the artifacts in the texture are meant to cause an Intensity drop that resembles the dust deposition effect. Single grains are modelled as Gaussian bivariate distributions, approximating the brightness drop in the area where the grain sticks to the lens. The distribution is generated on a squared grid and acquires all the possible color values per channel (Table \ref{tab:camera_par}). The corners of the grid are removed to make the blending with the background image smoother. The covariance matrix of the distribution and the peak value are randomized to mimic different grain shapes.\\
    The amount of rendered grains is set empirically to obtain a non-negligible darkening effect on the image and a statistically significant population of faults. Figure \ref{fig:dust_grains} provides a few examples of dust grain textures that are rendered in the dataset.
    \begin{table}[h!]
        \centering
        \begin{tabular}{p{4cm}|c|c|c}
            \textbf{Parameter}           &   \textbf{Effect}    & \textbf{Lower Bound}   & \textbf{Upper Bound} \\
            \hline
            Number of Dust Grains per Image & - & 30 & 100 \\
            Max of the Gaussian Distribution & Grain darkening & 100 & 200 \\
            Covariance of the Gaussian Distribution, $\sigma_{xx}$ & Grain Shape & 3 & 6 \\
            Covariance of the Gaussian Distribution, $\sigma_{yy}$ & Grain Shape & 3 & 6 \\
            Covariance of the Gaussian Distribution, $\sigma_{xy}=\sigma_{yx}$ & Grain Shape & 0 & 0 \\
        \end{tabular}
        \caption{Dust on Optics Fault Parameters.}
        \label{tab:dust_par}
    \end{table}

    % \begin{table}[h!]
    %     \centering
    %     \begin{tabular}{c|c|c|c|c}
    %          \textbf{Parameter} & \textbf{Grain 1} & \textbf{Grain 2} & \textbf{Grain 3} & \textbf{Grain 4} \\
    %          \hline
    %          Max of the Gaussian Distribution & 100 & 133.33 & 166.67 & 200 \\
    %          $\sigma_{xx}$ & 3 & 4 & 5 & 6 \\
    %          $\sigma_{yy}$ & 6 & 5 & 4 & 3 \\
    %          $\sigma_{xy}$ & 0 & 0 & 0 & 0 \\
    %          $\sigma_{yx}$ & 0 & 0 & 0 & 0 \\
    %      \end{tabular}
    %     \caption{Parameters of the Dust Grains at Figure \ref{fig:dust_grains}}
    %     \label{tab:dustgrains_params}
    % \end{table}    

    \subsubsection*{Broken Pixels}
    As for the Dust on Optics, the simulation of defect pixels makes use of pre-defined randomized textures to inject the faults onto the simulated image. Recalling the discussion about broken pixels at Section \ref{subs:cons_fail}, these defects can be related to issues of the sensor (i.e. impinging radiation) or the readout electronics. The two categories can occur both in the same image as they are uncorrelated. Instead, faults in CCD and CMOS sensors implies mutually-exclusive realizations of defects pixels, as the architecture of the sensor is fixed. 
    
\begin{figure}[h!]
        \centering
        \includegraphics[width=0.3\linewidth]{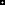}
        \caption{Sample Texture of a Single Broken Pixel.}
        \label{fig:singlebp_texture}
    \end{figure}

    The simulation of the single broken pixels does not change per sensor type considered, as the effect of a radiation impinging the detector causes a shift in the current output for both (Figure \ref{fig:bp_response}). Therefore, the texture used to simulate defect pixels is the same for CCD and CMOS sensors (Figure \ref{fig:singlebp_texture}).

    The number of simulated defects in the pixel depends on the specific radiation environment in which the mission operates. For an interplanetary orbit as the one of the \textit{Astrone KI} mission scenario, Solar Particles and Galactic Cosmic Rays (GCR) will strike the spacecraft undisturbed due to the lack of atmosphere and magnetic field.  However, the lifetime of \textit{Astrone KI} is not fixed yet but it is a crucial variable to determine the number of defective events that are expected to occur on the camera detector. As a simulation strategy, the number of broken pixels is set empirically to obtain statistically significant samples in the dataset, without making considerations on the influencing variables. This approach is deemed appropriate if the dataset is employed in a classification or segmentation task, as the one targeted by the present work.
    
    In order to simulate the dependency of a defect pixel response from the Exposure Time and ISO (Figure \ref{fig:bp_response} and \ref{fig:pixelresponse_ISO}), the texture at Figure \ref{fig:singlebp_texture} is injected with a randomized Intensity value (i.e. brightness). Indeed, on the same detector it is possible to find defect pixels with differently shifted outputs, based on the power transmitted by the impinging GCR. Different shifts will result in various brightness levels as they are readout by the electronics during the image acquisition. The neighbourhood pixels affected by the defect, well visible in Figure \ref{fig:singlebp_texture}, are introduced to keep into account demosaicing effects. The neighbourhood Intensity channel value is computed by linearly interpolating the Intensity of the defect pixel and the adjacent working one on the opposite side.
    
\begin{figure}[h!]
        \centering
        {
            \includegraphics[width=0.45\textwidth]{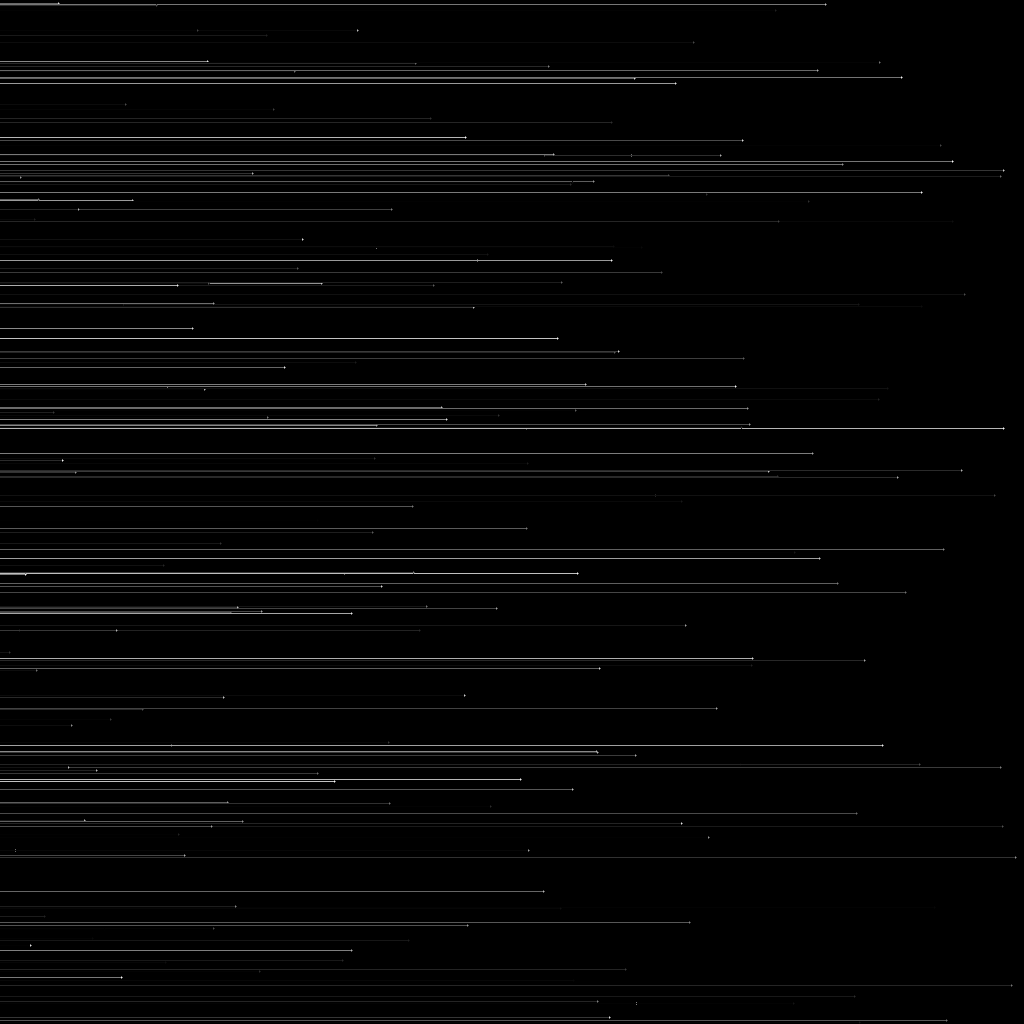}
        }
        {
            \includegraphics[width=0.45\textwidth]{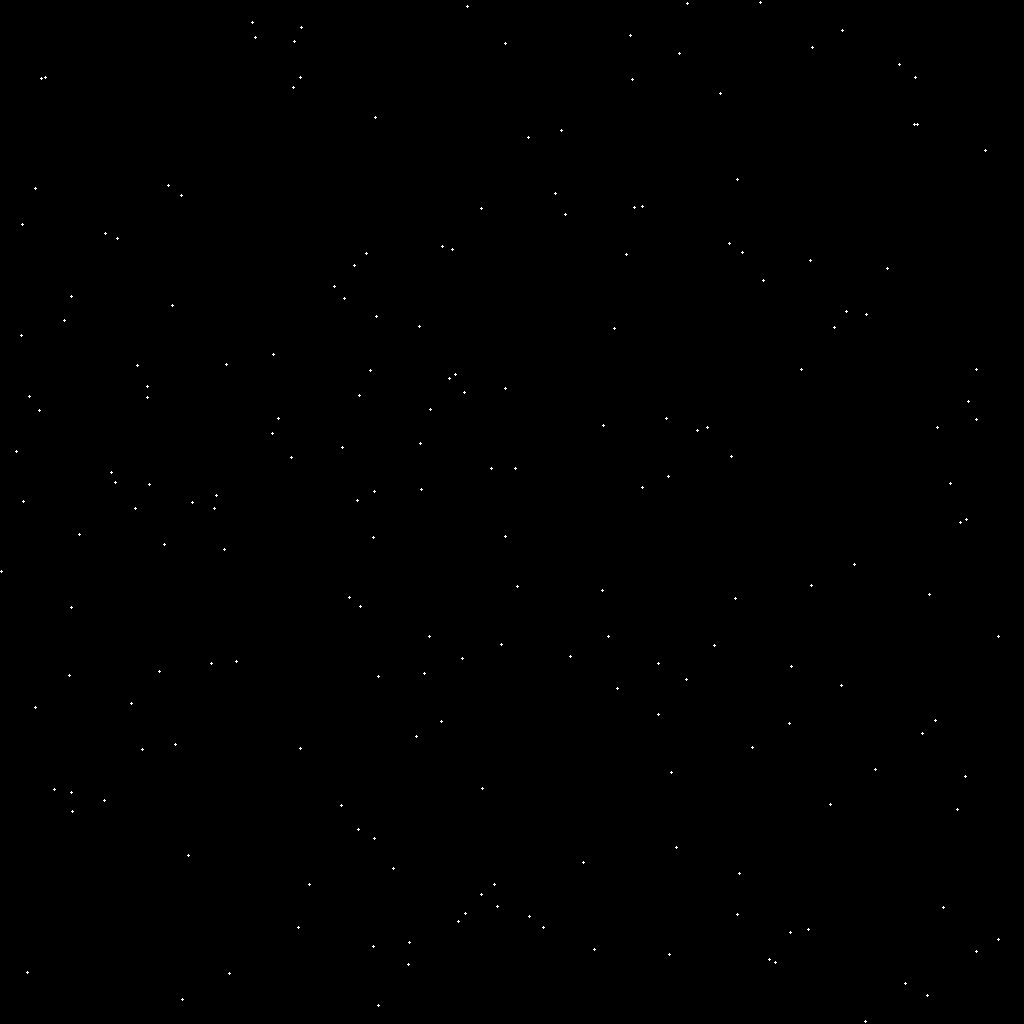}
        }
        % \hfill
        % {
        %     \includegraphics[width=0.22\textwidth]{ccd2.png}
        % }
        % \hfill
        % {
        %     \includegraphics[width=0.22\textwidth]{ccd3.png}
        % }
        \caption{Sample Textures of CCD (left) and CMOS (right) Broken Pixels. Textures represented in black background for visibility reasons.}
        \label{fig:ccd_cmos_bp_ex}
    \end{figure}

\begin{figure}[h!]
        \centering
        {
            \includegraphics[width=0.45\textwidth]{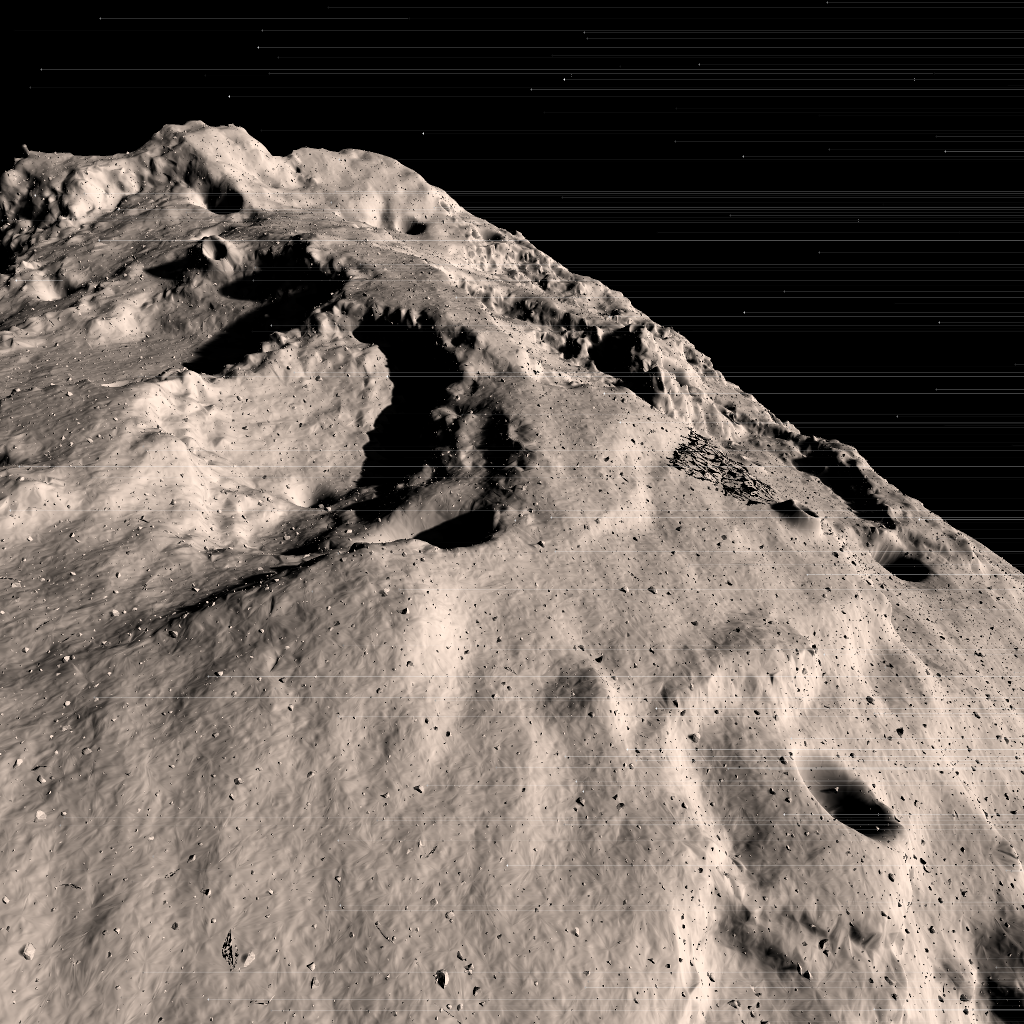}
        }
        {
            \includegraphics[width=0.45\textwidth]{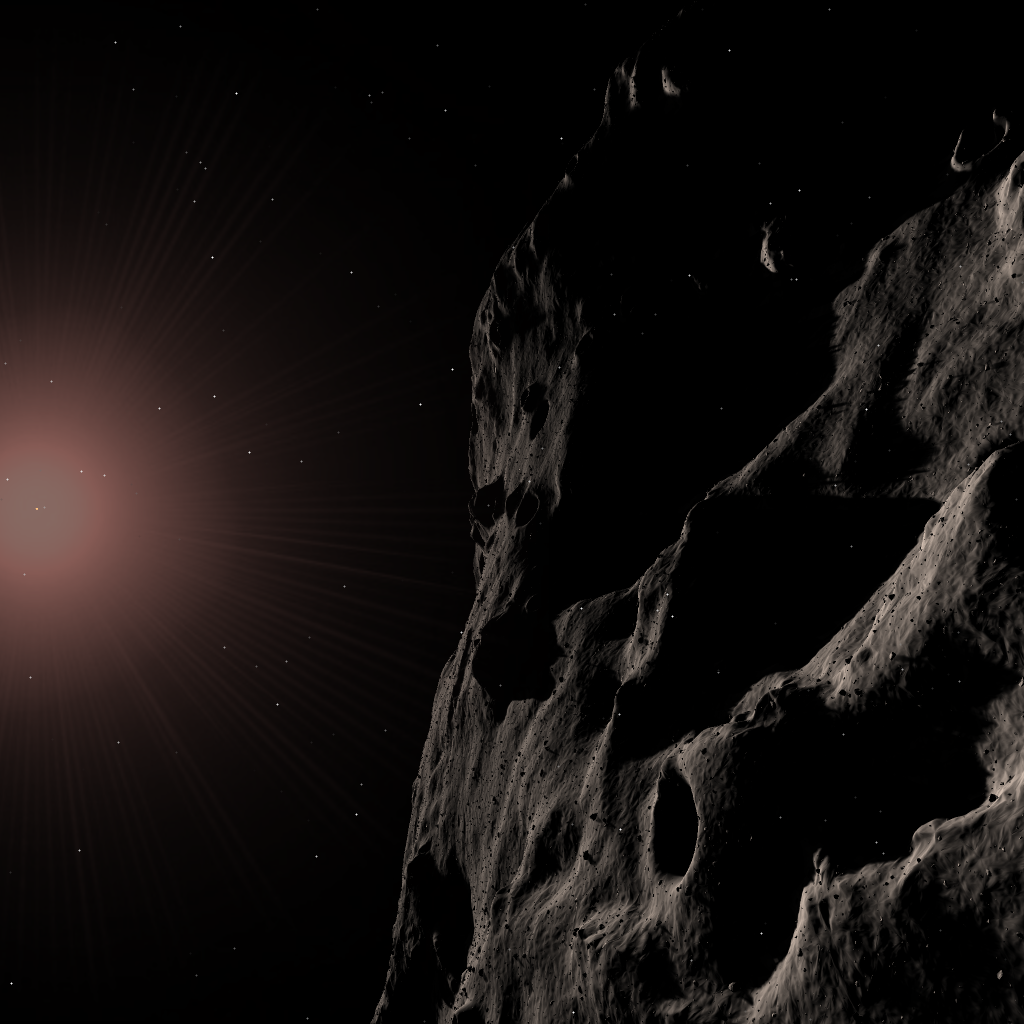}
        }
        \caption{Sample Images with CCD (left) and CMOS (right) Broken Pixels.}
        \label{fig:ccd_cmos_ex}
    \end{figure}

    As already stated in Section \ref{subs:cons_fail}, broken lines subtended to the respective defect pixels (in CCD detectors) derive their brightness from the respective pixel's offset. In order to simulate this effect, the lines are injected at a randomized level of Intensity between 0 (hot pixel without offset) and 0.4 $\times$ Intensity of the generator hot pixel. The direction of the lines in the final image is related to the readout mechanism position with respect to the surface of the detector. This means on one side that lines can have four different directions (top, bottom, right, left), on the other side that the direction is fixed within the same image. In other words, as the electronics is fixed by construction of the camera, all the generated lines will head towards the same direction of readout. In the simulation, a flag in introduced per image to randomize this feature. Figure \ref{fig:ccd_cmos_bp_ex} and \ref{fig:ccd_cmos_ex} respectively provide an example of injected texture and full rendered image, including faults from both kinds of sensors considered.

\begin{figure}[h!]
        \centering
        {
            \includegraphics[width=0.45\textwidth]{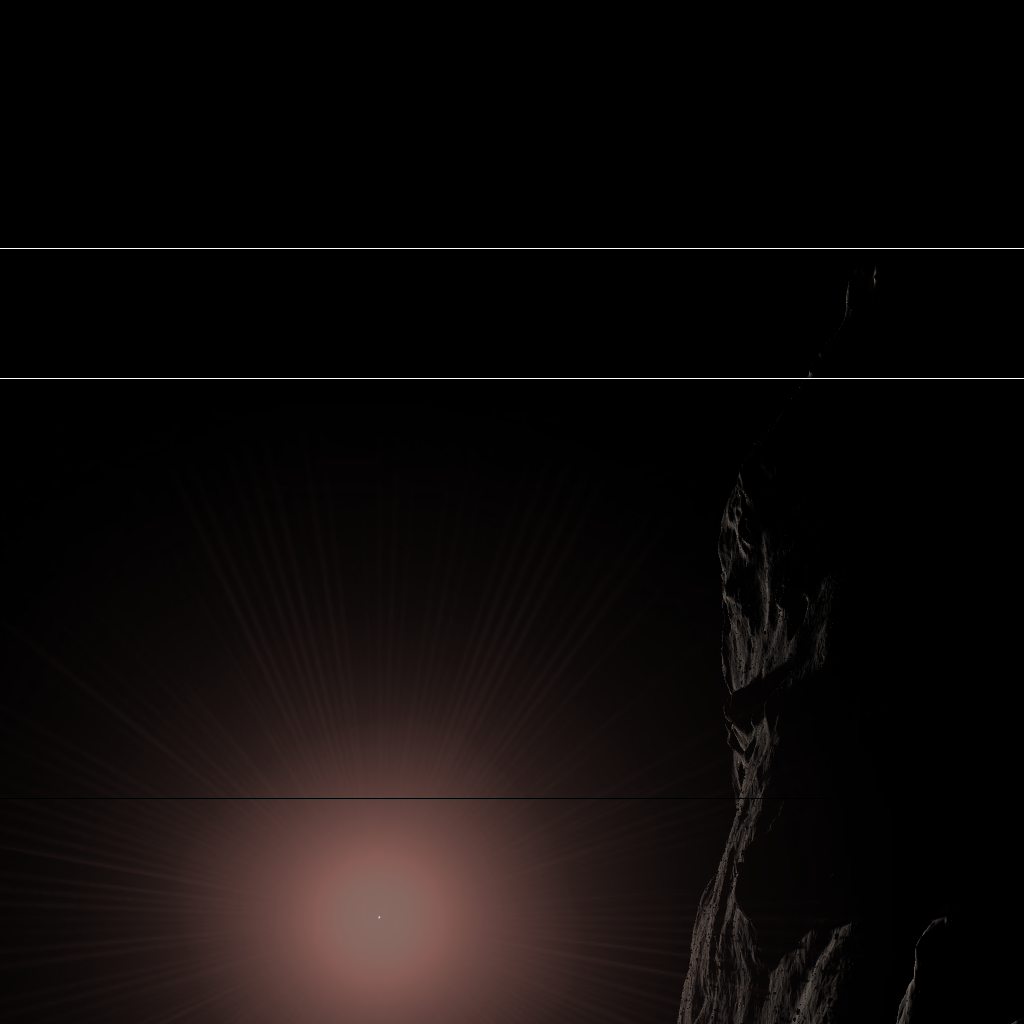}
        }
        {
            \includegraphics[width=0.45\textwidth]{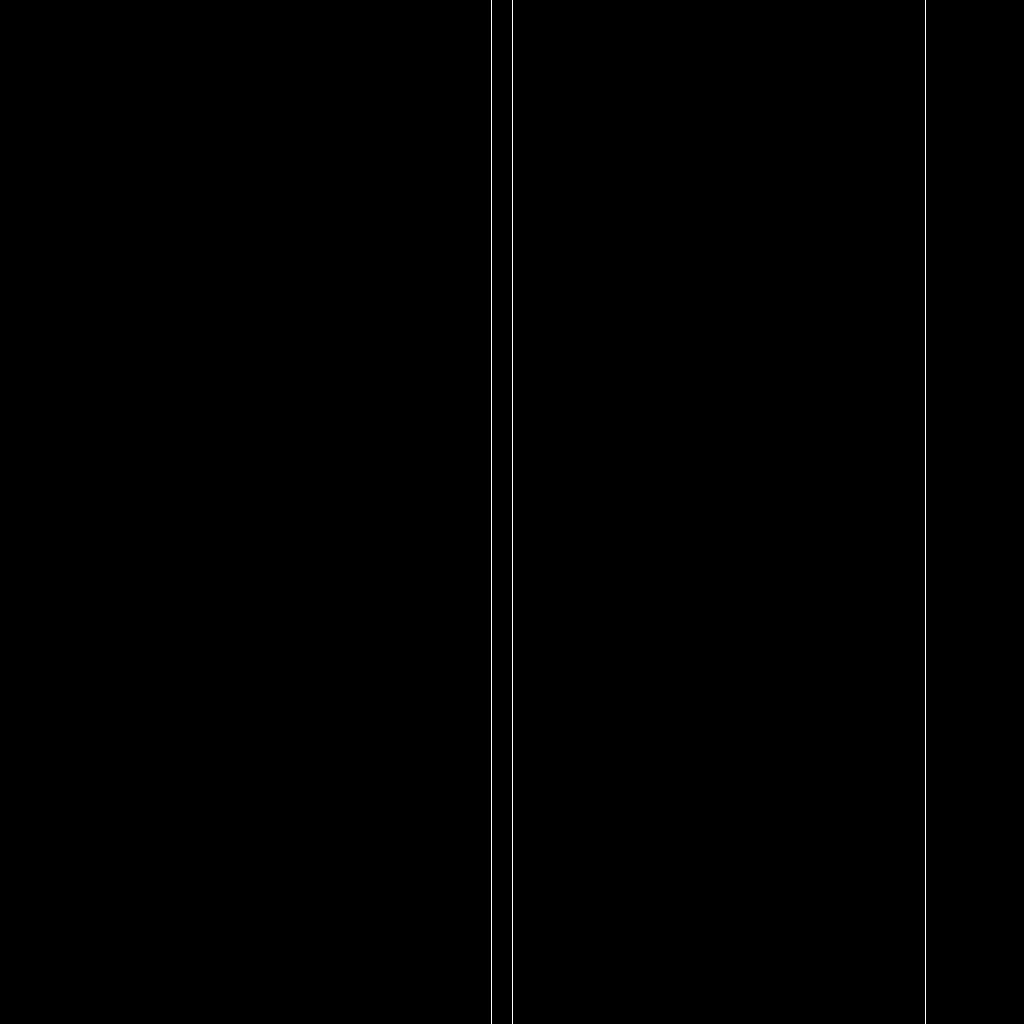}
        }
        \caption{Sample Images with Lines of Broken Pixels.}
        \label{fig:brokenlines_ex}
    \end{figure}

    Concerning broken lines deriving from faults of the electronics, the number of possible lines is empirically set to mimic the occurrence of such an issue in realistic onboard electronics. This occurrence is related to contingent factors and a robust and expensive design can be useful to reduce it to an acceptable level. The broken lines are simulated with randomized direction and Intensity values (Figure \ref{fig:brokenlines_ex}). The direction can be vertical or horizontal, depending on the readout components position with respect to the detector. To simulate different readout settings, a flag is introduced to command the direction of the lines. This flag also takes into account whether CCD or CMOS broken pixels are injected and adapts the injection of the lines accordingly (Section \ref{subs:cons_fail}). Finally, the Intensity value of the pixels composing the broken lines is randomized to assume values 0 or 255, providing the two cases of dead or saturated pixels. 


    Hence, the derived taxonomy of faults includes CCD and CMOS broken pixels (Figure \ref{fig:ccd_cmos_ex}), broken lines (Figure \ref{fig:brokenlines_ex} and mixed cases where both faults occur simultaneously.

    Finally, Table \ref{tab:bp_lines_par} provides simulation parameters for the Broken Pixels and Lines fault and their respective values.
    \begin{table}[h!]
        \centering
        \begin{tabular}{p{1.5cm}|p{5.5cm}|c|c|p{2cm}}
            \multirow{2}{*}{\textbf{Type}} & \multirow{2}{*}{\textbf{Parameter}}           & \textbf{Lower}   & \textbf{Upper}  & \multirow{2}{*}{\textbf{Categories}} \\
            &                  &  \textbf{Bound} & \textbf{Bound} &  \\
            \hline
             & Number of Broken Pixels per Image & 10 & 150 & - \\
            \textbf{Broken} & Brightness of the Pixel & 0 & 255 & - \\
            \textbf{Pixels} & Brightness of the Subtended Line (CCD) & 0 & 65 & - \\
            & Subtended Line Direction (CCD) & - & - & Up, Down, Right, Left \\
            \hline
            \textbf{Broken} & Number of Broken Lines & 1 & 5 & - \\
            \textbf{Lines} & Entity of the Broken Lines & - & - & Black, White \\
        \end{tabular}
        \caption{Broken Pixels and Lines fault Parameters, Dimensionless.}
        \label{tab:bp_lines_par}
    \end{table}

    \subsubsection*{Straylight}
    The simulation strategy for straylight adopts the approach described in \citet{king2001game} of rendering pre-defined textures onto the final image. Although it can be considered outdated in the state of the art, this methodology finds its motivation in the scope of the presented dataset. In fact, by not relying on the definition of a lenses setup, while freely injecting random textures, it aims to reach representativeness of a variety of different scenarios and camera setups that can occur in future space missions employing visual sensors. Clearly, the downside of this approach with respect to the physically-based ones described in \citet{keshmirian2008physically}, \citet{hullin2011physically}, \citet{lee2013practical} is that the latter yield generally more realistic images. In this work, realism has a lower priority than being able to target a wide panorama of potential lens geometries, but the integration of physically-based models is an interesting path of future work. 
 
\begin{figure}[h!]
    \centering
        {
            \includegraphics[width=0.17\textwidth]{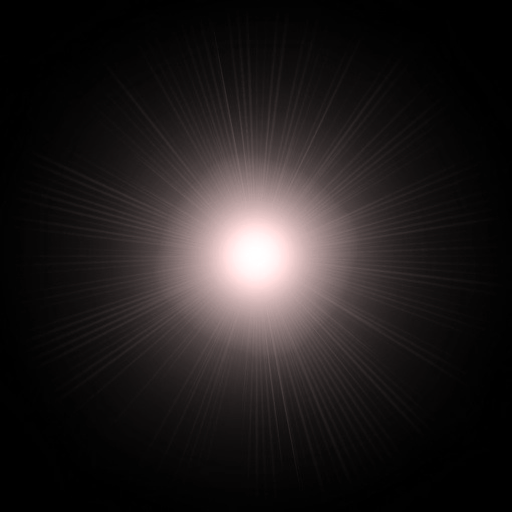}
        }
        {
            \includegraphics[width=0.17\textwidth]{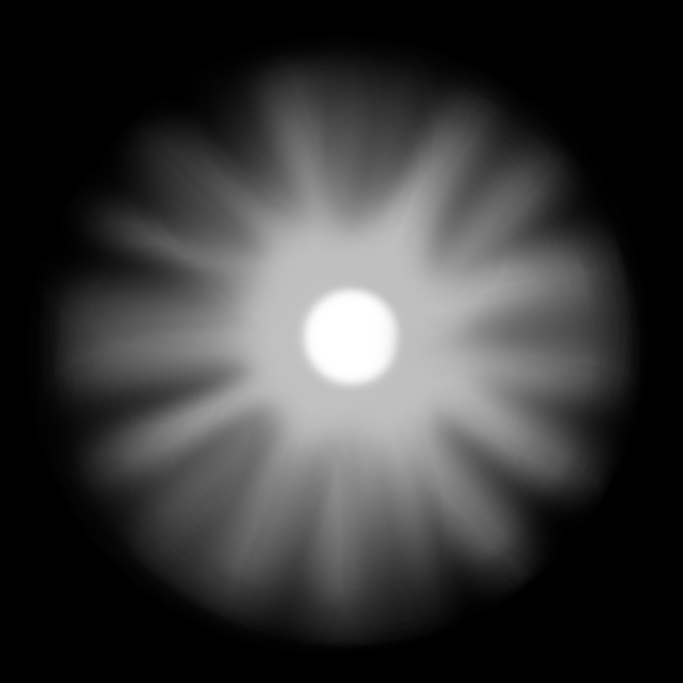}
        }
        \caption{Injected Glares Textures.}
        \label{fig:sl_textures_glares}
    \end{figure}

\begin{figure}[h!]
    \centering
        {
            \includegraphics[width=0.17\textwidth]{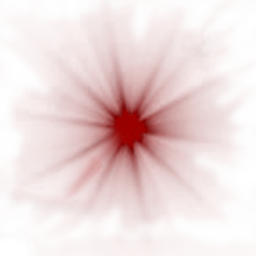}
        }
        \hfill
        {
            \includegraphics[width=0.17\textwidth]{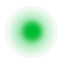}
        }
        \hfill
        {
            \includegraphics[width=0.17\textwidth]{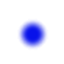}
        }
        \hfill
        {
            \includegraphics[width=0.17\textwidth]{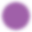}
        }
        \hfill
        {
            \includegraphics[width=0.17\textwidth]{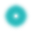}
        }
    
    \vspace{1em}
    
        {
            \includegraphics[width=0.17\textwidth]{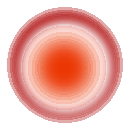}
        }
        \hfill
        {
            \includegraphics[width=0.17\textwidth]{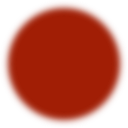}
        }
        \hfill
        {
            \includegraphics[width=0.17\textwidth]{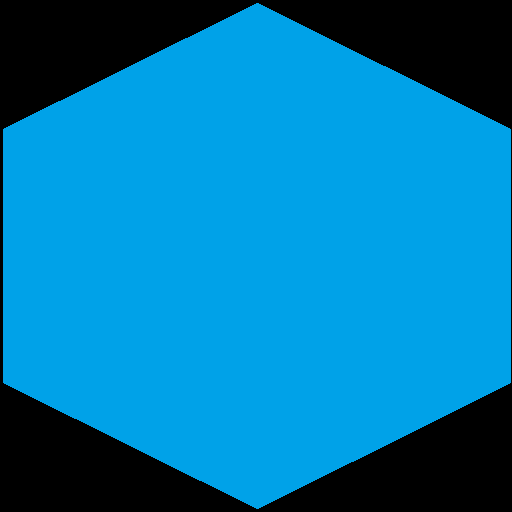}
        }
        \hfill
        {
            \includegraphics[width=0.17\textwidth]{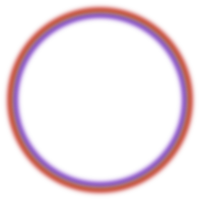}
        }
        \hfill
        {
            \includegraphics[width=0.17\textwidth]{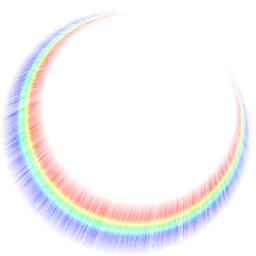}
        }
        
        \caption{Injected Flares Textures, divided in \textit{Close} (top) and \textit{Far} (bottom).}
        \label{fig:sl_textures_flares}
    \end{figure}

    The textures employed for glares and flares injection are presented in Figure \ref{fig:sl_textures_glares} and \ref{fig:sl_textures_flares}. Although it has been stated that the random textures injection is the baseline for the present approach, a specific adjustment was made to enhance the realism of the simulation. Based on the graphical results of \citet{hullin2011physically}, \citet{keshmirian2008physically}, \citet{lee2013practical} (Figure \ref{fig:phys_flares} from \citet{hullin2011physically}), it is assumed that the rendered flares follow an increasing-size pattern as the artifacts occur further from the light source. Moreover, textures closer to the light source present \textit{denser} layouts (i.e. orbs), while the further they are, the more diverse shapes they acquire (i.e. rings, polygons). To reproduce this behaviour in the present work without assuming a specific lens layout, the textures are split in two groups,\textit{close} and \textit{far}. The separation between the two groups is always kept along the parameters randomization process described later in this paragraph. 
    
    According to \citet{king2001game}, the rendered flares shall lie on the line connecting the center of the image and the light source, i.e. the Sun. A 2D reference frame is defined along this line, where the light source lies at 0 and the center of the image lies at 1 (dimensionless units). The flares positions can also assume values outside the (0, 1) interval, considering that for positions $\gg1$ and $\ll0$ flares can fall outside the image frame, sacrificing the visibility (i.e. resulting in no noteworthy impact during processing). Additionally, the size and brightness of the flares are expressed in dimensionless units and are essential for regulating the respective characteristics. The values for size and brightness serve as multiplication factors to derive the final size and brightness after rendering, based on the original texture values.
    
    The simulation approach includes as a first step a random selection of which flares to inject among the ones in Figure \ref{fig:sl_textures_flares}. Then, a random value of position, radius and brightness is assigned to each of the selected flares, that are finally  rendered onto the image. While randomizing the position parameters, it is ensured that textures from the \textit{close} group (top part of Figure \ref{fig:sl_textures_flares}) are rendered closer to the light source than the ones coming from the \textit{far} group (bottom part of Figure \ref{fig:sl_textures_flares}). Boundaries for the flares parameters are listed in Table \ref{tab:sl_par}.
    \begin{table}[h!]
        \centering
        \begin{tabular}{c|c|c}
            \textbf{Parameter}                   & \textbf{Lower Bound}   & \textbf{Upper Bound} \\
            \hline
            Number of Injected Flares   & 1             & 10 \\
            Index of Injected Flares    & 1             & 10 \\
            Flares Position             & 0.5           & 2 \\
            Flares Radius               & 0.05          & 0.3 \\
            Flares Brightness           & 1.5           & 2.5 \\
        \end{tabular}
        \caption{Straylight Fault Parameters.}
        \label{tab:sl_par}
    \end{table}

\begin{figure}[h!]
    \centering
    {
        \includegraphics[width=0.45\textwidth]{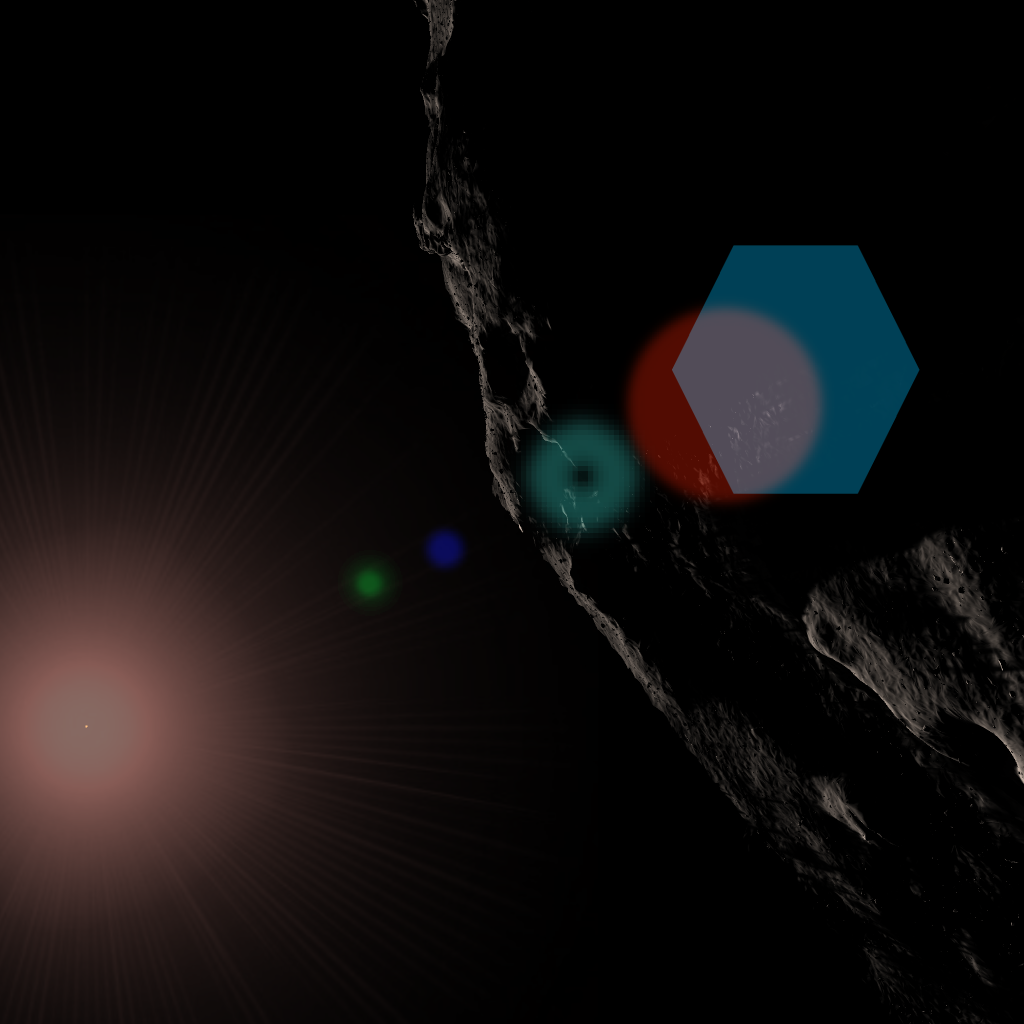}
    }
    {
        \includegraphics[width=0.45\textwidth]{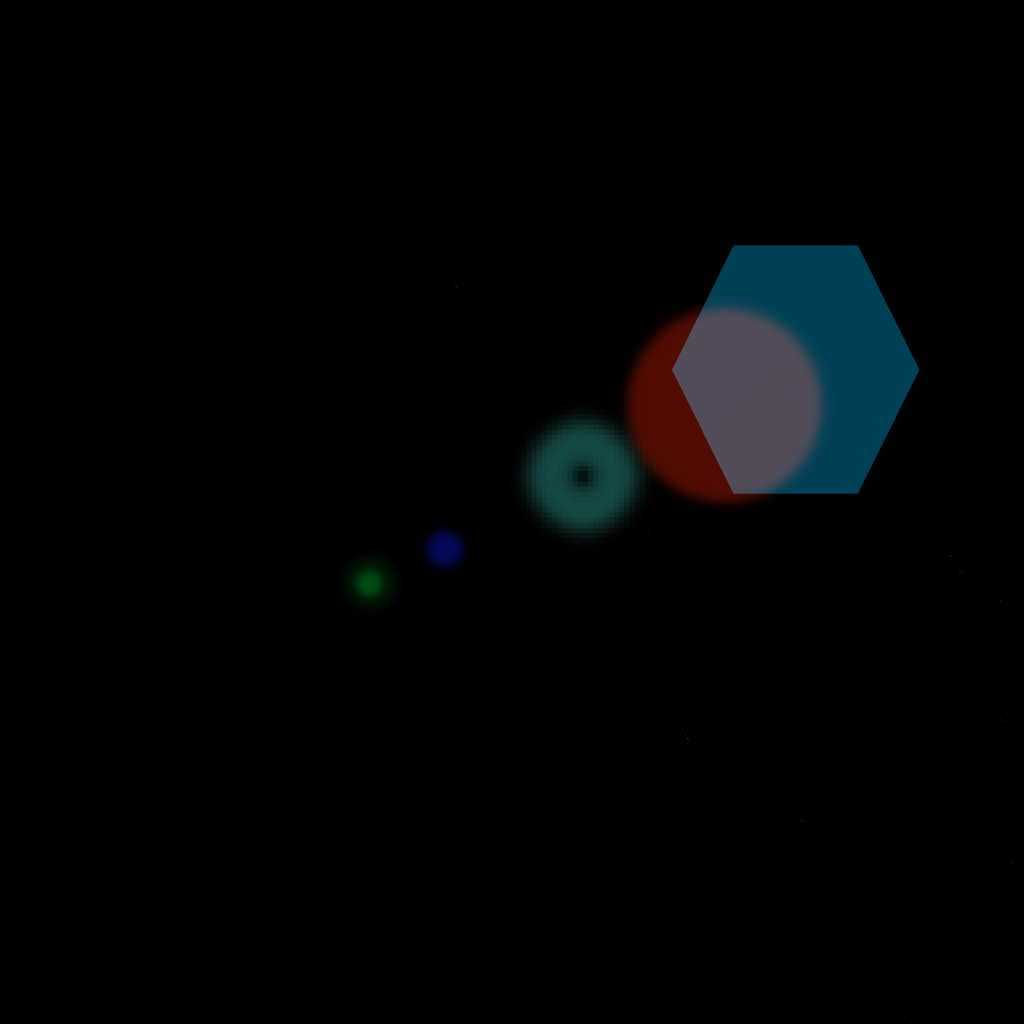}
    }
    \caption{Sample Image affected by Straylight (left) and the Injected Texture (right)}
    \label{fig:sl_ex1}
    \end{figure}

\begin{figure}[h!]
        \centering
        {
            \includegraphics[width=0.45\textwidth]{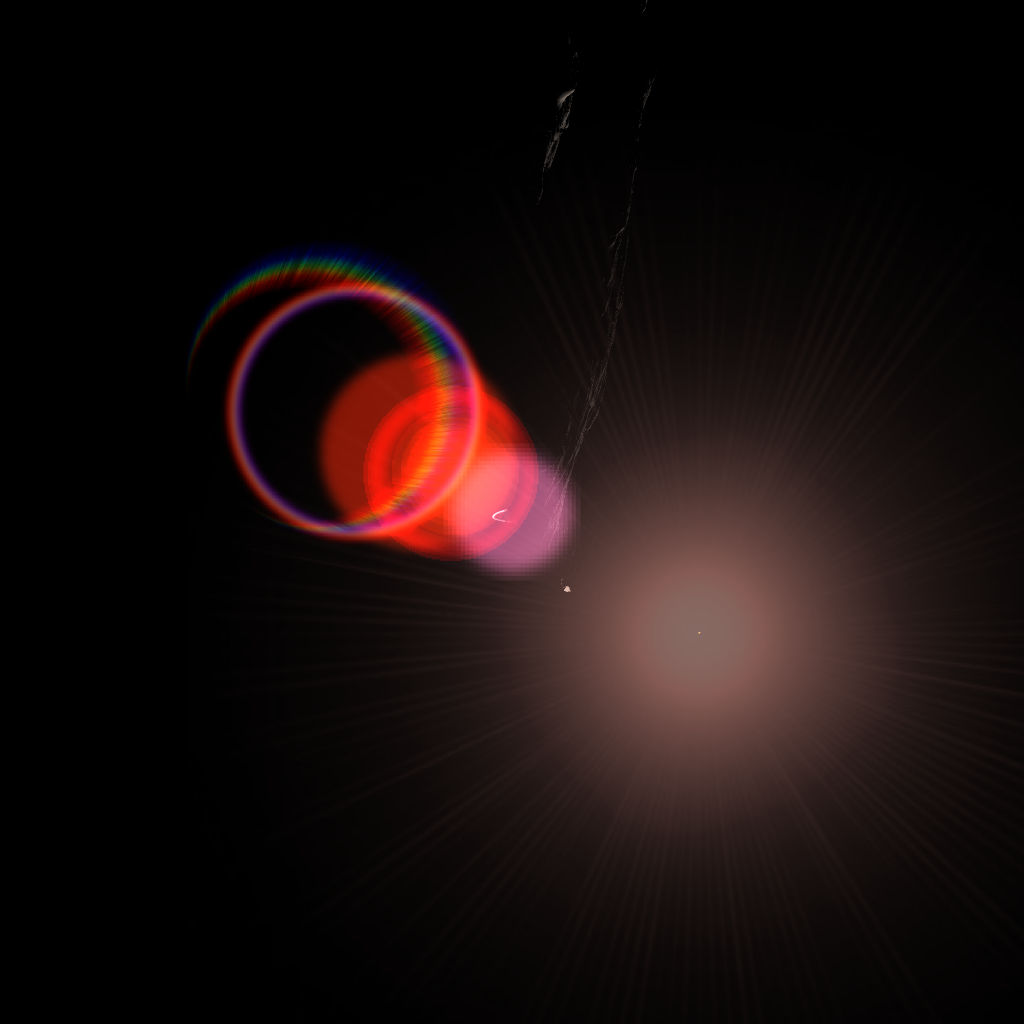}
        }
        {
            \includegraphics[width=0.45\textwidth]{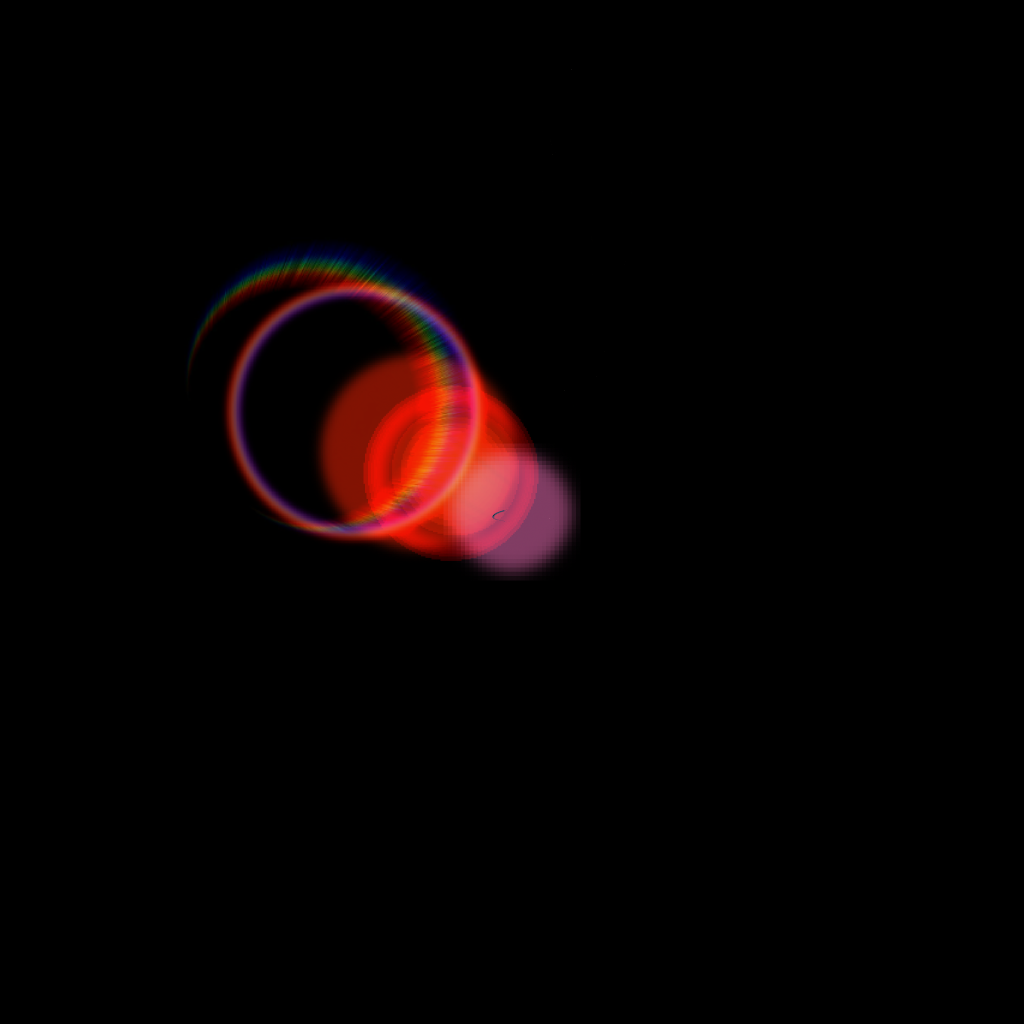}
        }
        \caption{Sample Image affected by Straylight (left) and the Injected Texture (right)}
        \label{fig:sl_ex2}
    \end{figure}

    Figure \ref{fig:sl_ex1} and \ref{fig:sl_ex2} provide examples of images affected by straylight.

    \subsubsection*{Vignetting}
\begin{figure}[h!]
        \centering
        {
        \includegraphics[width=0.45\textwidth]{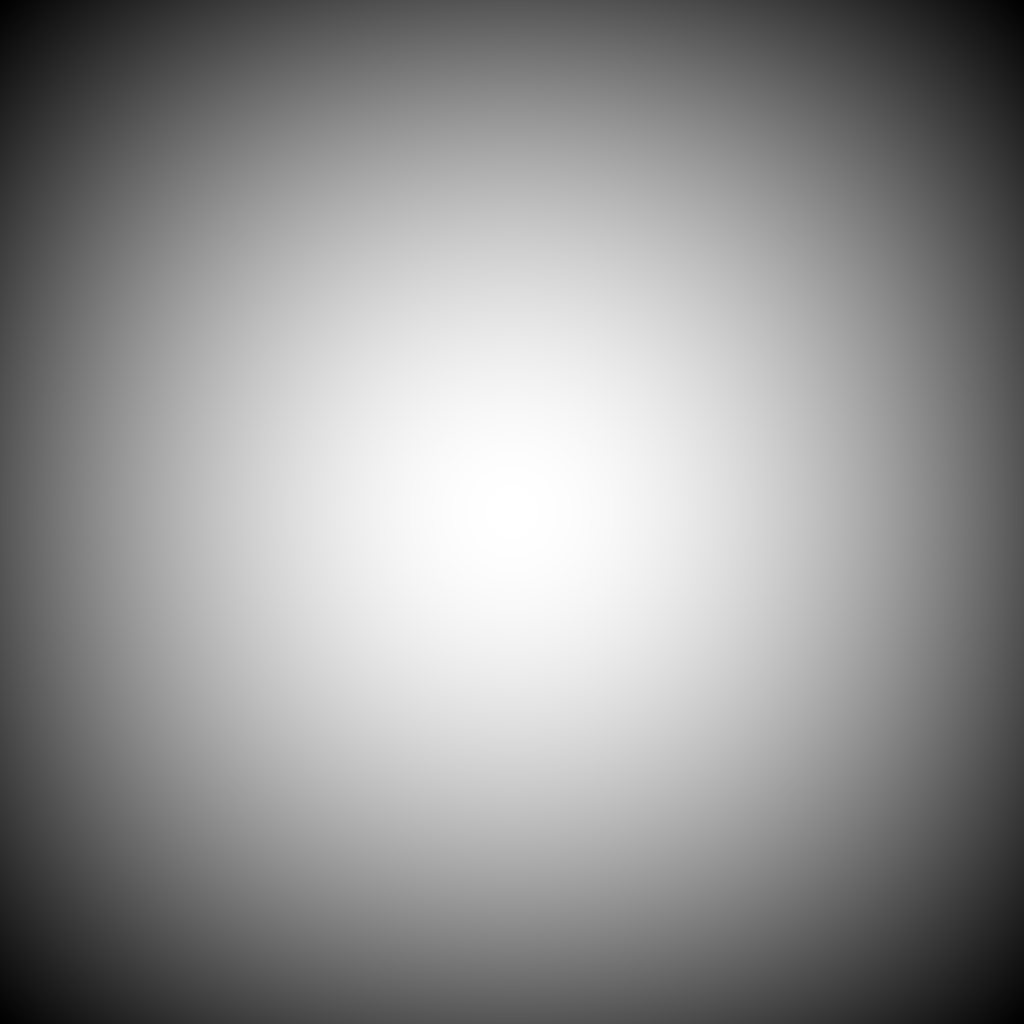}
        }
        {
        \includegraphics[width=0.45\textwidth]{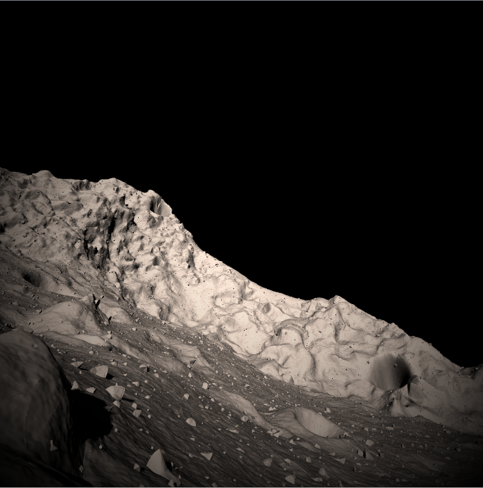}
        }
        \caption{Sample Texture (left) and Sample Image for Vignetting Effect (right). On-axis illuminance is set to 255.}
        \label{fig:vign_text_ex}
    \end{figure}

    The vignetting effect is simulated in the present work using a monochromatic texture generated to provide the brightness decay effect described in Equation \ref{eq:cos4_law} in Section \ref{subs:cons_fail}. The texture has the size of the acquired image and it assigns to each pixel an Intensity value based on the aforementioned law. The angle in Equation \ref{eq:cos4_law} is computed using the pixel coordinates, while the intensity of the vignetting effect is given by setting the on-axis illuminance ($E_{0}$) to an arbitrary value. This value is randomized to provide diverse darkening scenarios. An example of the texture applied to the sample image and the produced effect is provided in Figure \ref{fig:vign_text_ex}.
    
    Table \ref{tab:vign_par} provides boundaries and explanation of Vignetting parameters employed in the dataset generation.
    \begin{table}[h!]
        \centering
        \begin{tabular}{c|c|c|c}
            \textbf{Parameter}           &   \textbf{Effect}    & \textbf{Lower Bound}   & \textbf{Upper Bound} \\
            \hline
            On-axis Illuminance & Darkening effect & 105 & 255 \\
            \end{tabular}
        \caption{Vignetting Fault Parameters}
        \label{tab:vign_par}
    \end{table}

    \subsubsection*{Optics Degradation}
    In the present work, the only effect of optics degradation considered is \textit{blurriness} uniformly affecting the camera image. A more detailed modelling can be implemented as future work. The simulation of blurriness makes use of a Gaussian Filter convolved with the image. The Gaussian Filter is parametrized by a matrix of adjustable size including values drawn from a Gaussian distribution. The size of the filter varies the number of pixels affected and consequently the intensity of the blur (Table \ref{tab:blur_par}).
    \begin{table}[h!]
        \centering
        \begin{tabular}{c|c|c|c}
            \textbf{Parameter}           &   \textbf{Effect}    & \textbf{Lower Bound}   & \textbf{Upper Bound} \\
            \hline
            Gaussian Filter Size (pixels) & Blur Intensity & 3 & 17 \\
            \end{tabular}
        \caption{Optics Degradation Fault Parameters.}
        \label{tab:blur_par}
    \end{table}

\begin{figure}[h!]
        \centering
        \includegraphics[width=0.45\textwidth]{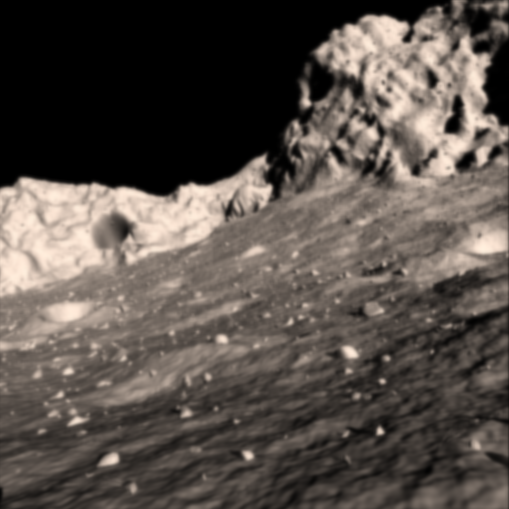}

        \caption{Sample Image for Optics Degradation Fault}
        \label{fig:blur_ex}
    \end{figure}

    An example of how the blurriness realizes in a rendered image is provided in Figure \ref{fig:blur_ex}. 

\subsection{Labelling}
The labelling of a dataset is strongly related to the targeted task while training and testing the AI on the dataset itself. In this work, supervised learning is assumed, but no specific task. Examples of tasks suitable to VBN comprise but are not limited to direct image classification and segmentation. Both of them aim at the recognition of specific artifacts of the observed terrain to be used for e.g. pose estimation or trajectory planning in the downstream navigation algorithm. The difference resides in the fact that classification assigns a class to a specific image, while segmentation marks groups of pixels belonging to the same class within one image. Examples of datasets for classification in the frame of VBN include \citet{MERdataset} and \citet{MSLdataset}, which provide images of specific Mars landmarks such as rocks, boulders, Sun, sky, rover parts, etc. The size of the images in these datasets shall be sufficiently small to include one specific landmark in each image, avoiding cases of equivocal classification between two different classes. 

In this work, binary segmentation masks are generated per fault, meaning that each image presents a number of labelling masks equal to the number of fault classes in the dataset. The masks can be employed to perform both classification or segmentation, including binary or multi-class classification and binary or multi-class segmentation. Also, it is possible to address the faults incrementally, according to the needs, exploiting the separation of the masks. A binary mask labelling all the faults in the same image is also included, in case the targeted task is segmenting out failed areas instead than recognizing the specific fault.
The labelling strategy is straightforward for all the fault cases, except for straylight. Indeed, for those fault cases characterized by a pre-computed texture (Dust on Optics, Broken Pixels and Lines, Vignetting) or an index commanding the injection (Optics Degradation), both the texture and the index can be used as a label themselves. For the straylight, the simulation infrastructure does not output the texture of the injected flares, thus it needs \textit{a posteriori} retrieval. To do this, a post-processing step is carried out after the main one, including only the images facing the Sun (Table \ref{tab:bound_var}), to obtain the same images without rendered flares. Subtracting the new images from the ones obtained in the nominal generation procedure allows to isolate the flares from the rest of the image and obtain the desired masks.

\begin{figure}[h!]
    \centering
        \includegraphics[width=0.8\textwidth]{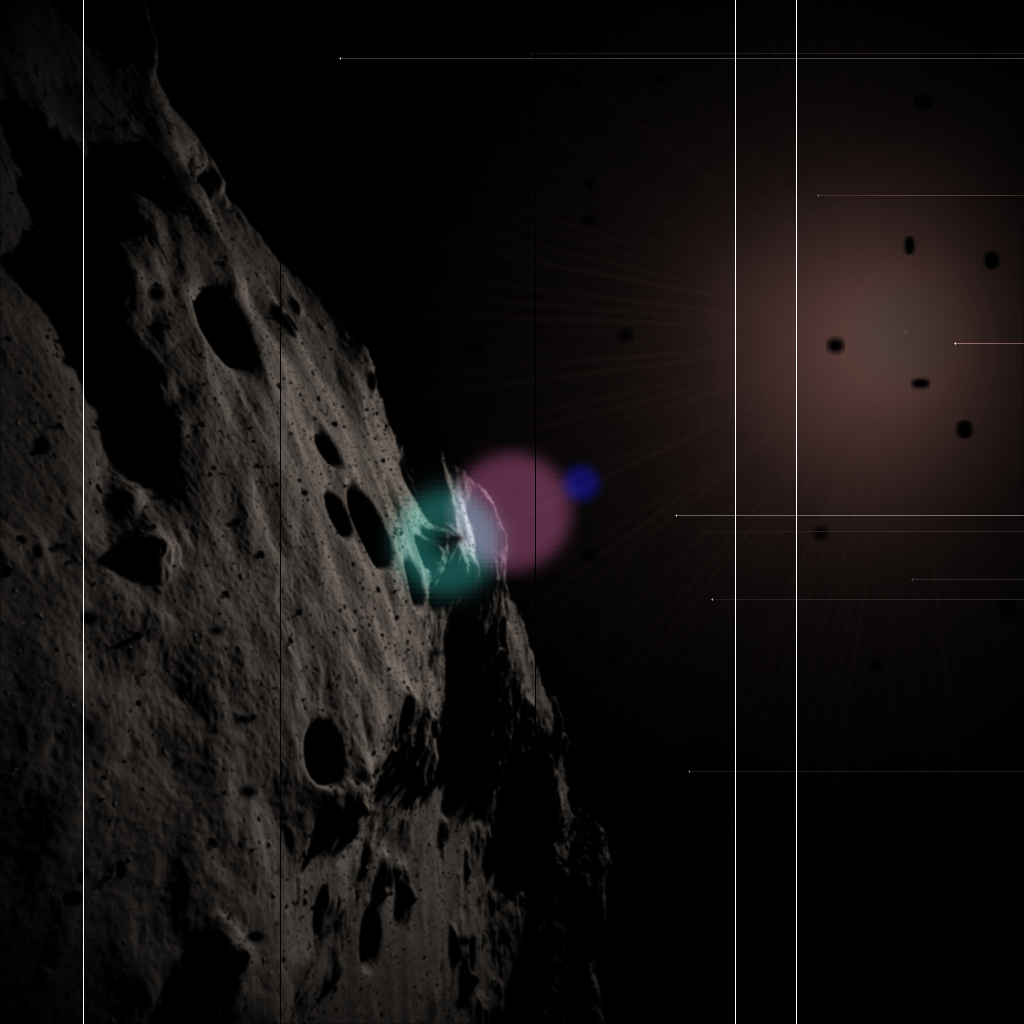}
    \caption{Sample Image Full Rendering}
    \label{fig:fullex_4mask}
\end{figure}

Figure \ref{fig:fullex_4mask} shows an example of a full rendered image. The parameters of the injected faults for this picture are provided in Table \ref{tab:param_4mask}. The image counterpart with no rendered straylight and the rest of the masks are provided in \ref{app:B}. 
\begin{table}[h!]
    \centering
    \begin{tabularx}{\linewidth}{X|X|X}
         & \textbf{Parameter}           & \textbf{Value} \\
        \hline 
        \textbf{Dust on Optics} & Number of Dust Grains per Image &  50\\
        \hline
        \multirow{3}{*}{\textbf{Broken Pixels}} & Sensor Type & CCD \\
        & Number of Broken Pixels per Image & 9\\
        & Subtended Line Direction (CCD) & right \\
        \hline
        \multirow{2}{*}{\textbf{Broken Lines}} & Number of Broken Lines & 5 \\
        & Entity of the Broken Lines & 3 White, 2 Black \\
        \hline
        \multirow{5}{*}{\textbf{Straylight}} & Number of Injected Flares   & 3\\
        & Index of Injected Flares    & 3, 4, 5 \\
        & Flares Position             & 0.83, 1.00, 1.17 \\
        & Flares Radius               & 0.11, 0.13, 0.16 \\
        & Flares Brightness           & 1.50, 1.50, 1.50 \\
        \hline
        \textbf{Vignetting} & On-axis Illuminance & 255 \\
        \hline
        \textbf{Optics Degradation} & Gaussian Filter Size (Pixels) & 3 \\
    \end{tabularx}
    \caption{Parameters of the Injected faults in Figure \ref{fig:fullex_4mask}}
    \label{tab:param_4mask}
\end{table}

% \section{Dataset Analytics}
% Give results.\\
% Total number of images, size, memory...\\
% Percentage of samples per class, binary and multi-class case\\
% Update based on read papers

\section{Conclusion and Future Work} \label{sec:concl_fw}

In this study, a novel methodology for the simulation of faults occurring in typical Vision-based Navigation systems is presented. The research focuses on common issues that can be experienced in the scenario of interplanetary exploration such as the \textit{Astrone KI} one, providing a comprehensive analysis of the faults occurrence and effects in image data. Nevertheless, generalization is expected to any mission employing visual sensors with diverse needs. 

A novel method is proposed to simulate the identified faults, employing cross-domain studies and methodologies tailored to the specific space application. The study proposes a level of detail of the simulation that is deemed sufficient for the data to be representative of the relative fault scenario, but it could easily be expanded in the future according to the needs.

Finally, a dataset is derived simulating together 5000 images with injected fault cases. The dataset simulates the environmental condition of the comet 67P/Churyumov-Gerasimenko, capturing a wide range of scenarios and illumination levels, marking a valuable resource for developing and testing AI algorithms in VBN systems. This dataset sets a significant step in literature with its unique contribution, and aims at triggering the scientific research towards the development of reliable and safe autonomous system for spacecraft navigation.

Looking ahead, the natural follow-up of the present work is the employment of the dataset in a practical use-case of VBN where it can be used for training and testing an AI to perform anomaly detection. Such an idea can be logically applied to the \textit{Astrone KI} system, where an AI-based anomaly detection step would enhance the robustness of the downstream VBN pipeline, allowing it to work safely without fearing failed samples that can cause issues in the processing. Moreover, the authors encourage wide use of the dataset across various mission types, with the expectation that it will generalize effectively. To this purpose, Table \ref{tab:fail_missapp} presents an overview of different space mission classes, suggesting the applicability of the faults analysed in the present work.
\begin{table}[h!]
    \centering
    \caption{Applicability of Fault Cases based on Common Mission Objectives}
    \label{tab:fail_missapp}
    \begin{tabularx}{\textwidth}{c|c|XXXX}
         & & \multicolumn{4}{c}{\textbf{Classes}} \\
        \hline
        &  & \textbf{Earth Observation} & \textbf{Landing} & \textbf{SSSB Exploration} & \textbf{Interplanetary} \\
        \cline{1-6}
        \multirow{6}{*}{\begin{sideways} \textbf{Faults}\end{sideways}} 
         & \textbf{Dust on Optics} & N & Y & Y & N \\
         & \textbf{Broken Pixels} & N$^{*2}$ & N & Y & Y \\
         & \textbf{Broken Lines} & Y & Y* & Y & Y \\
         & \textbf{Straylight} & Y & N & Y & Y \\
         & \textbf{Vignetting} & Y & Y & Y & Y \\
         & \textbf{Optics Degradation} & N$^{*1}$ & N & Y & Y \\
         % \cline{2-7}
    \end{tabularx}
    \begin{tablenotes}
        \item[*] *: faults causing Broken Lines can also occur in the cruising phase
        \item[$^{*1}$] $^{*1}$: Given that sufficiently high-quality parts, shielding and ultimately correction algorithms are employed
        \item[$^{*2}$] $^{*2}$: Carefully adapt the radiation environment
    \end{tablenotes}
\end{table}

\section*{Acknowledgements}

The results presented in this paper have been achieved by the project Astrone - Increasing the Mobility of Small Body Probes, which has received funding from the German Federal Ministry for Economic Affairs and Energy (BMWi) under funding number “50 RA 2130A”. The consortium consists of Airbus Defence and Space GmbH, Astos Solutions GmbH, Institute of Automation (Technische Universität Dresden) and Institute of Flight Mechanics and Controls (Universität Stuttgart). Responsibility of the publication contents is with the publishing author.

\appendix
\section{Reference Frames} \label{app:A}
Table \ref{tab:ref_frames} details the reference frames employed in the present study (Section \ref{sec:datagen}).
\begin{table}[h!]
    \centering
    \begin{tabular}{p{3.5cm}|p{3.8cm}|p{7cm}}
        \textbf{Reference Frame} & \textbf{Origin} & \textbf{Axes} \\
        \hline
        International Celestial, $\mathcal{J}$ & Solar System's Barycenter & \textbf{\textit{x}-axis:} The intersection of equatorial and ecliptic planes (vernal equinox) at 12:00 Terrestrial Time on 1 January 2000, 12:00:00, \newline \textbf{\textit{z}-axis:} Normal to the equator at 12:00 Terrestrial Time on 1 January 2000, 12:00:00 in the direction of the north pole, \newline \textbf{\textit{y}-axis:} Completes the right hand rule.\\
         & & \\
        Target-Body Centered Target-Body Fixed, $\mathcal{T}$ & CoM of the target body & \textbf{\textit{x}-axis:} Towards the axis with the least MoI, \newline \textbf{\textit{y}-axis:} Completes the right hand rule, \newline \textbf{\textit{z}-axis:} Towards the axis with the highest MoI. \\
         & & \\
        Local-Level, $\mathcal{L}$ & User-defined on the target surface & \textbf{\textit{z}-axis:} Aligned with the effective gravity vector, \newline \textbf{\textit{x}-axis:} Points in north direction and is perpendicular to the \textit{z}-axis, \newline \textbf{\textit{y}-axis:} Completes the right-hand system. \\
         & & \\
        Body-Fixed, $\mathcal{B}$ & CoM of the spacecraft & \textbf{\textit{z}-axis:} Downwards directed, perpendicular to the horizontal bottom panel, \newline \textbf{\textit{y}-axis:} Perpendicular to the \textit{z}-axis in the direction of one solar panel, \newline \textbf{\textit{x}-axis:} Completes the right hand rule.\\
         & & \\
        Camera-Fixed, $\mathcal{C}$ & CoM of the spacecraft & \textbf{\textit{z}-axis:} Same as the Body Fixed frame, \newline \textbf{\textit{y}-axis:} Same as the Body Fixed frame, \newline \textbf{\textit{x}-axis:} Same as the Body Fixed frame.\\
    \end{tabular}
    \caption{Reference Frames Summary Table}
    \label{tab:ref_frames}
\end{table}

\clearpage

\section{Sample Masks} \label{app:B}

\begin{figure}[h!]
    \centering
    \includegraphics[width=\textwidth]{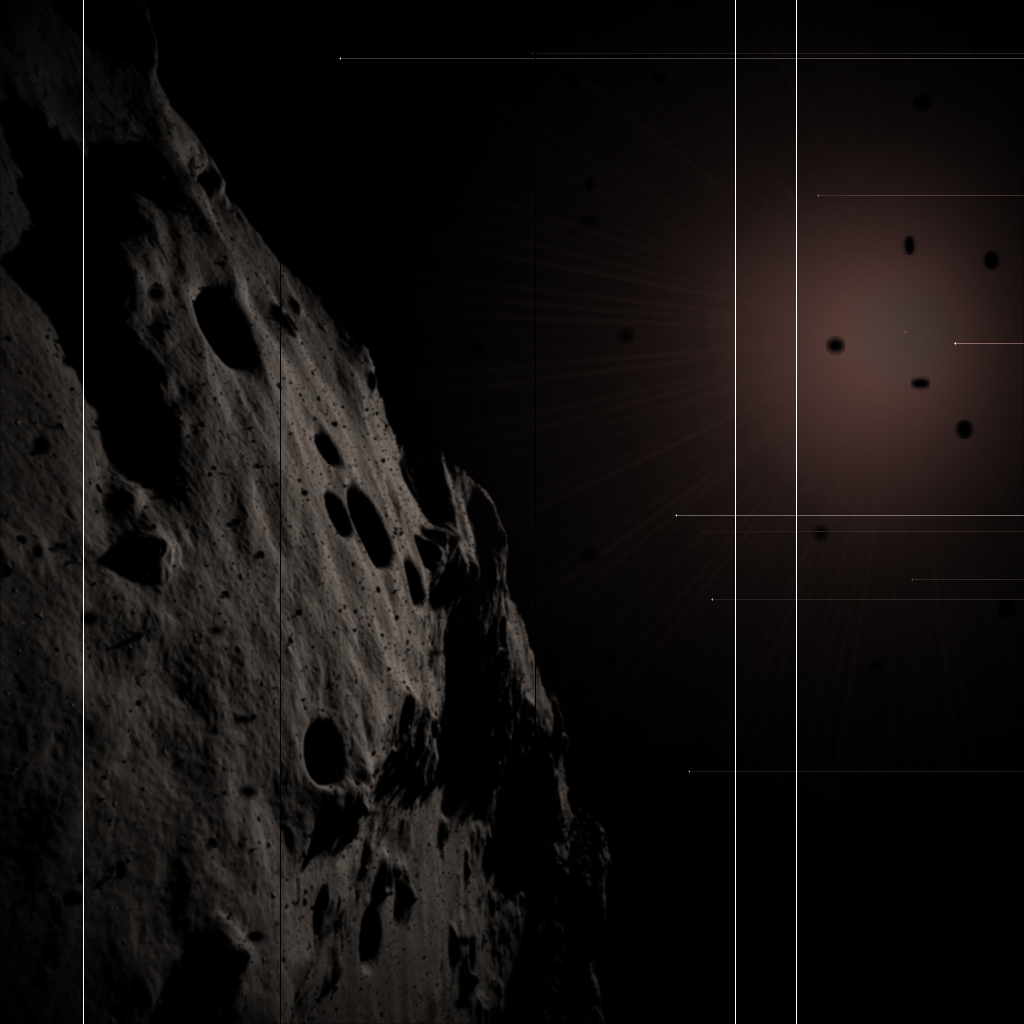}
    \caption{Figure \ref{fig:fullex_4mask} without Rendered Straylight}    \label{fig:fullexnoSL_4mask}
\end{figure}

\begin{figure}[h!]
    \centering
    \includegraphics[width=\textwidth]{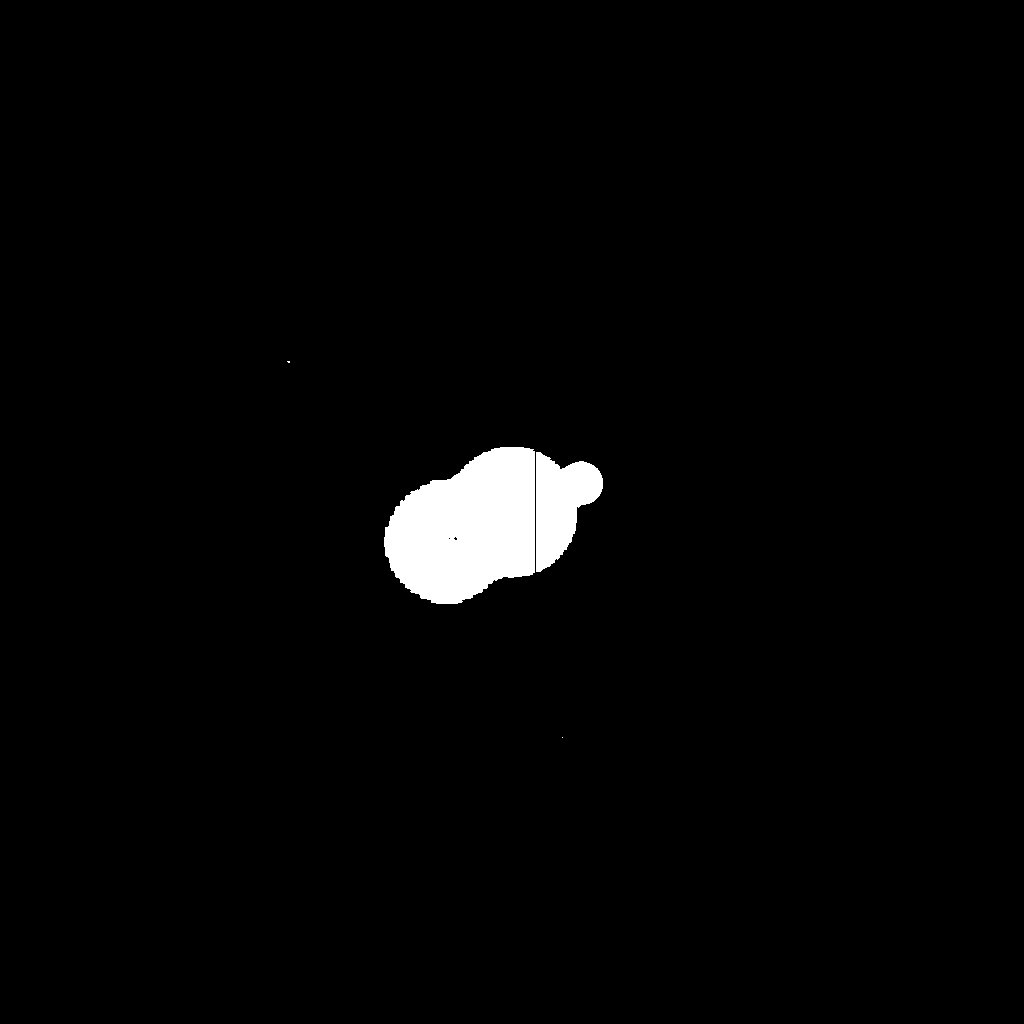}
    \caption{Straylight Mask. Note that the black line intersecting the flare textures appears black because it is a Broken Line, so it is not part of the Straylight label.}
    \label{fig:SL_4mask}
\end{figure}

\begin{figure}[h!]
    \centering
    \includegraphics[width=\textwidth]{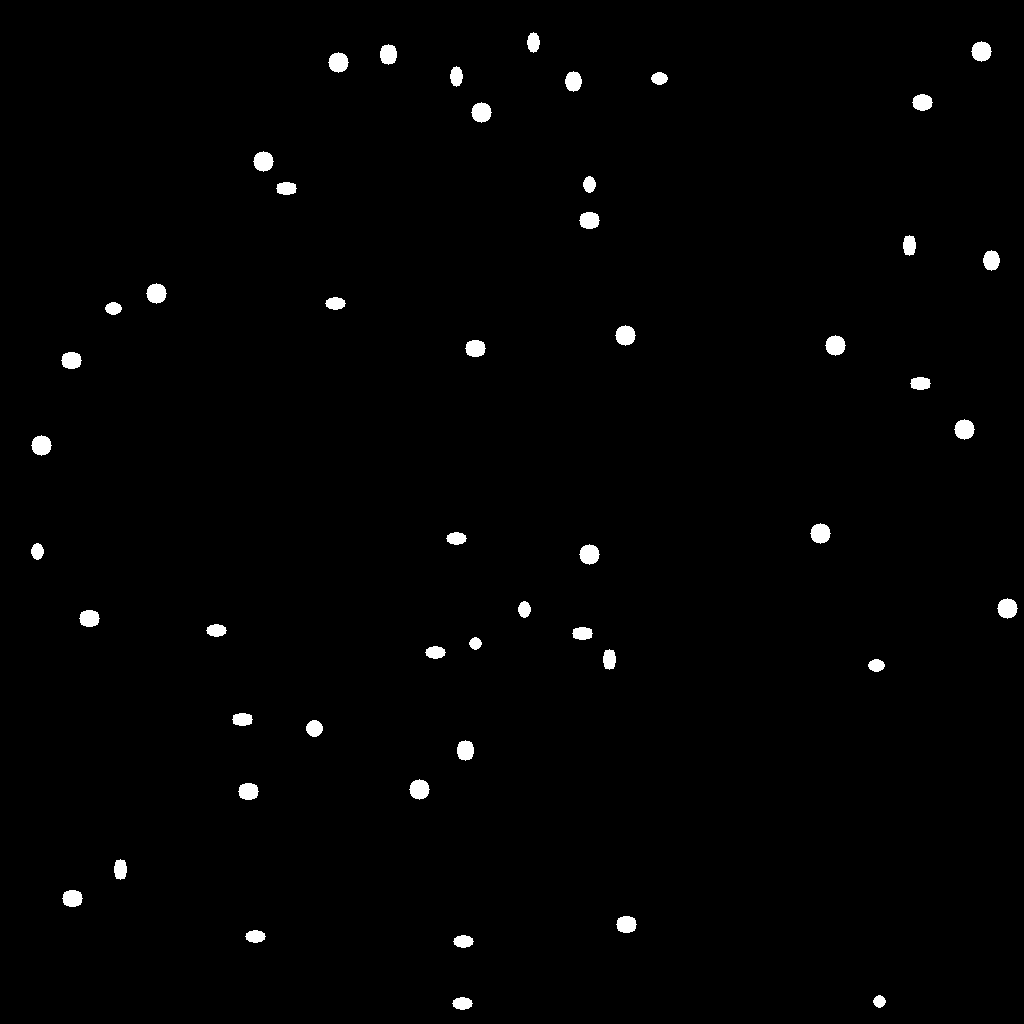}
    \caption{Dust on Optics Mask}
    \label{fig:dust_4mask}
\end{figure}

\begin{figure}[h!]
    \centering
        \includegraphics[width=\textwidth]{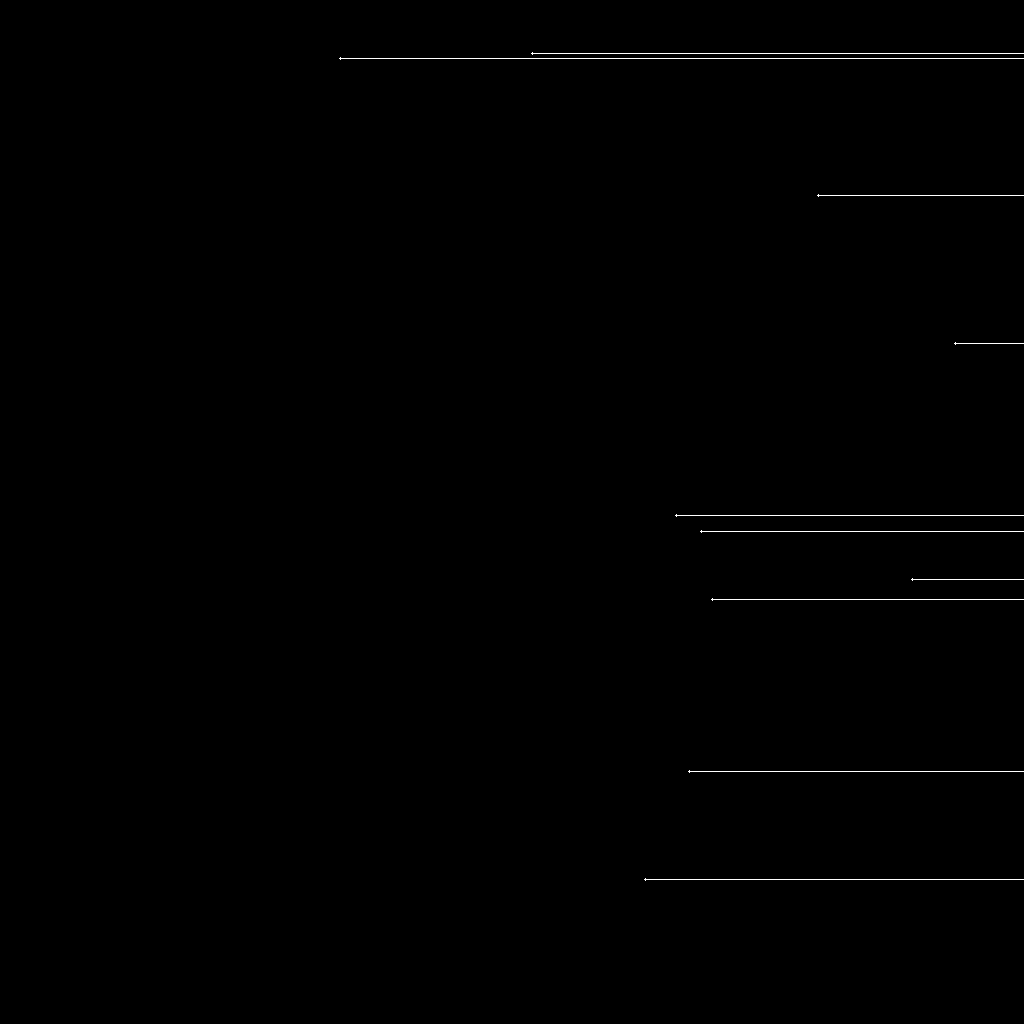}
    \caption{Broken Pixels Mask}
    \label{fig:BP_4mask}
\end{figure}

\begin{figure}[h!]
    \centering
        \includegraphics[width=\textwidth]{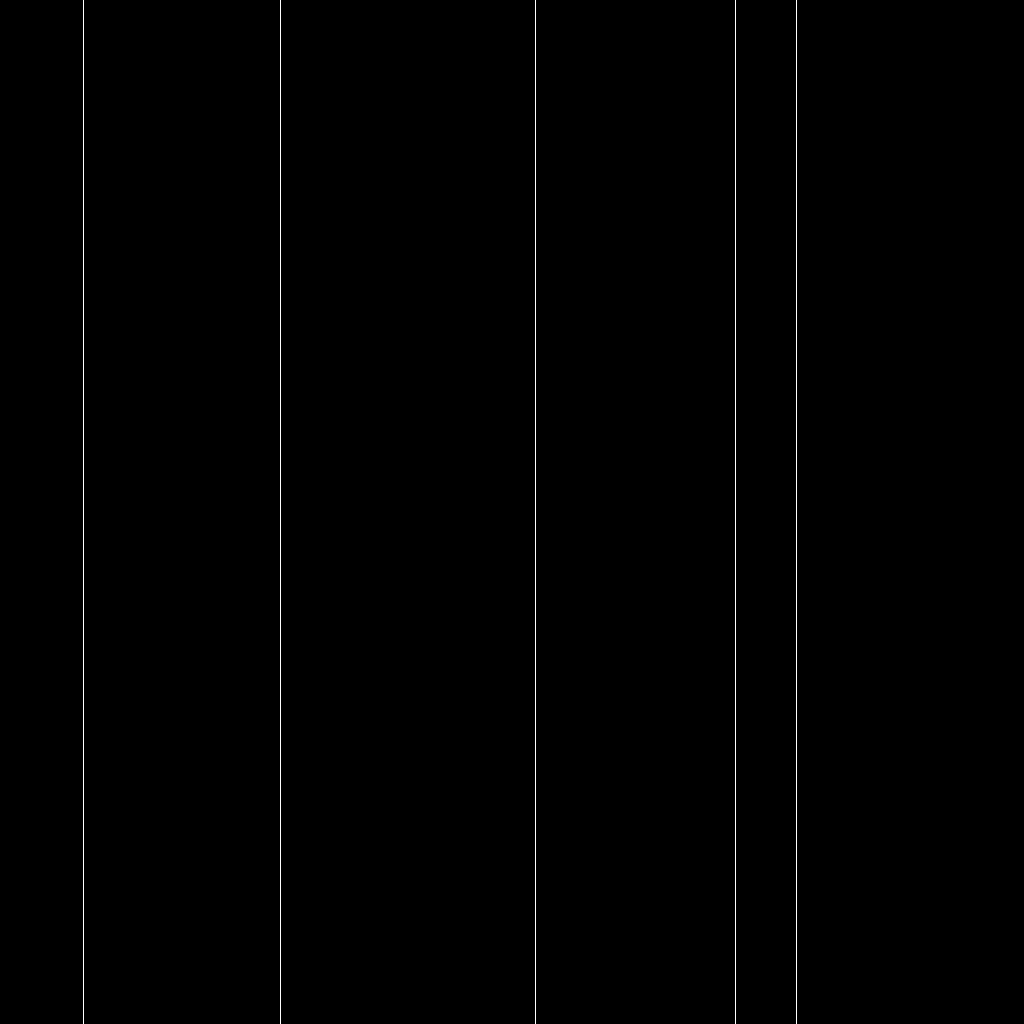}
    \caption{Broken Lines Mask}
    \label{fig:BL_4mask}
\end{figure}

Based on Figure \ref{fig:fullex_4mask}, Figure \ref{fig:fullexnoSL_4mask} present its counterpart without straylight. The retrieved straylight mask is shown in Figure \ref{fig:SL_4mask}. \ref{fig:dust_4mask} shows the Dust on Optics mask. Finally, Figure \ref{fig:BP_4mask} and Figure \ref{fig:BL_4mask} provides the masks associated to Broken Pixels and Broken Lines.

\begin{figure}[h!]
    \centering
    \includegraphics[width=\textwidth]{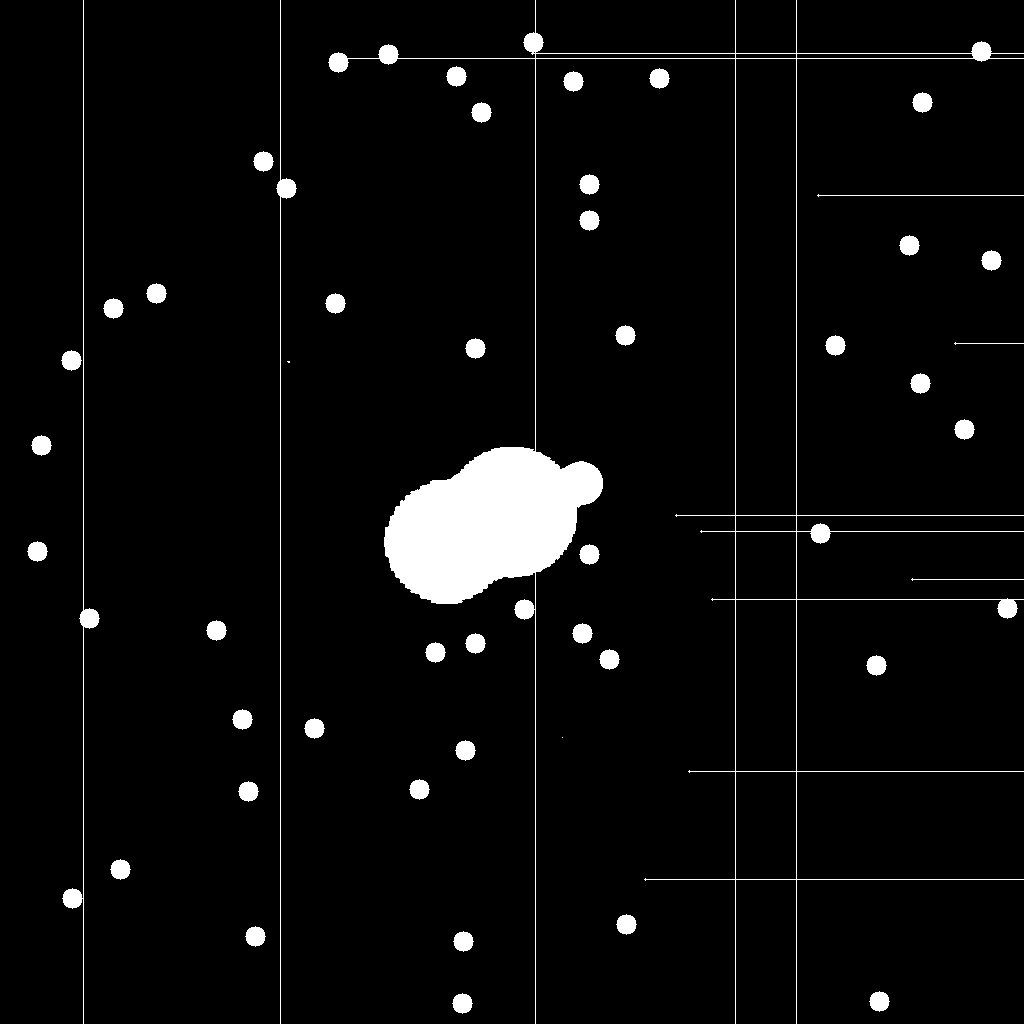}
    \caption{Sample Mask Full faults}
    \label{fig:fullmask_4mask}
\end{figure}

\clearpage








\bibliographystyle{elsarticle-num-names} 
\bibliography{sample}
\end{document}